\documentclass[10pt,twocolumn,letterpaper]{article}

\usepackage{cvpr}
\usepackage{times}
\usepackage{epsfig}
\usepackage{graphicx}
\usepackage{amsmath}
\usepackage{amssymb}
\usepackage{subfigure}

\usepackage[pagebackref=true,breaklinks=true,letterpaper=true,colorlinks,bookmarks=false]{hyperref}

\cvprfinalcopy 

\def\cvprPaperID{388} 

\ifcvprfinal\fi
\begin{document}

\title{Do We Need Binary Features for 3D Reconstruction?}

\author{
Bin Fan$^{\ast}$ \qquad Qingqun Kong$^{\ast}$ \qquad Wei Sui$^{\ast}$ \qquad Zhiheng Wang$^{\ddagger}$ \qquad Xinchao Wang$^{\dagger}$ \\
\qquad Shiming Xiang$^{\ast}$ \qquad Chunhong Pan$^{\ast}$ \qquad Pascal Fua$^{\dagger}$ \\
$^{\ast}$ Institute of Automation, Chinese Academy of Sciences\\
    $^{\ddagger}$ School of Computer Science and Technique, Henan Polytechnic University\\
    $^{\dagger}$ CVLab, EPFL 
}
    
\maketitle

\begin{abstract}
   Binary features have been incrementally popular in the past few years due to their low memory footprints and the efficient computation of Hamming distance between binary descriptors. They have been shown with promising results on some real time applications, e.g., SLAM, where the matching operations are relative few. However, in computer vision, there are many applications such as 3D reconstruction requiring lots of matching operations between local features. Therefore, a natural question is that is the binary feature still a promising solution to this kind of applications? To get the answer, this paper conducts a comparative study of binary features and their matching methods on the context of 3D reconstruction in a recently proposed large scale mutliview stereo dataset. Our evaluations reveal that not all binary features are capable of this task. Most of them are inferior to the classical SIFT based method in terms of reconstruction accuracy and completeness with a not significant better computational performance.
\end{abstract}

\section{Introduction}
Matching local features across multiple images captured from different viewpoints and positions plays a fundamental role in image based 3D reconstruction~\cite{Snavely_SIGGRAPH06,Strecha_CVPR10,Frahm_10,Agarwal_11,Heinly_CVPR15}. Feature matching involves extracting local keypoints from images, constructing local descriptors for keypoints, and establishing point correspondences across different images according to distances of descriptors. SIFT~\cite{LOWE_IJCV04} has been a popular method for keypoint extraction and description in the past decade for the task of 3D reconstruction. It uses a 128-dimensional float point vector, which is known as SIFT descriptor, to represent the local information of a keypoint. Such a high dimensional float point representation results in a large memory footprint, limiting its potential in large scale and embedded applications. Meanwhile, computing Eculidean distances between SIFT descriptors is also time consuming when it has to compute lots of them. These disadvantages motivate researchers to put efforts on studying binary descriptors in recent years~\cite{Calonder_PAMI11,FREAK_CVPR12,Leutenegger_ICCV11,Rublee_ICCV11,Wang_BMVC13,Fan_TIP14,Trzcinski_PAMI15}.

Unlike float point descriptors, binary descriptors use a binary string to describe a keypoint. Due to the characteristic of binary string, storing a binary descriptor only requires 1/32 memory of that used by storing a float point descriptor of the same dimension. Another advantage of binary descriptor lies in its matching speed. Modern computer architecture has fully supported the computation of Hamming distance between binary descriptors by simple machine instructions. Therefore, computing the Hamming distance is usually 1-2 orders of magnitude faster than computing the corresponding Eculidean distance. Although the research of binary descriptors has been flourished in the past few years, they have not yet been widely used except for some light weight tasks, e.g., template based object detection~\cite{Calonder_PAMI11} and SLAM~\cite{ORB_SLAM}.

\begin{figure}[t]
	\centering \subfigure[]{
		\includegraphics[width=0.23\textwidth]{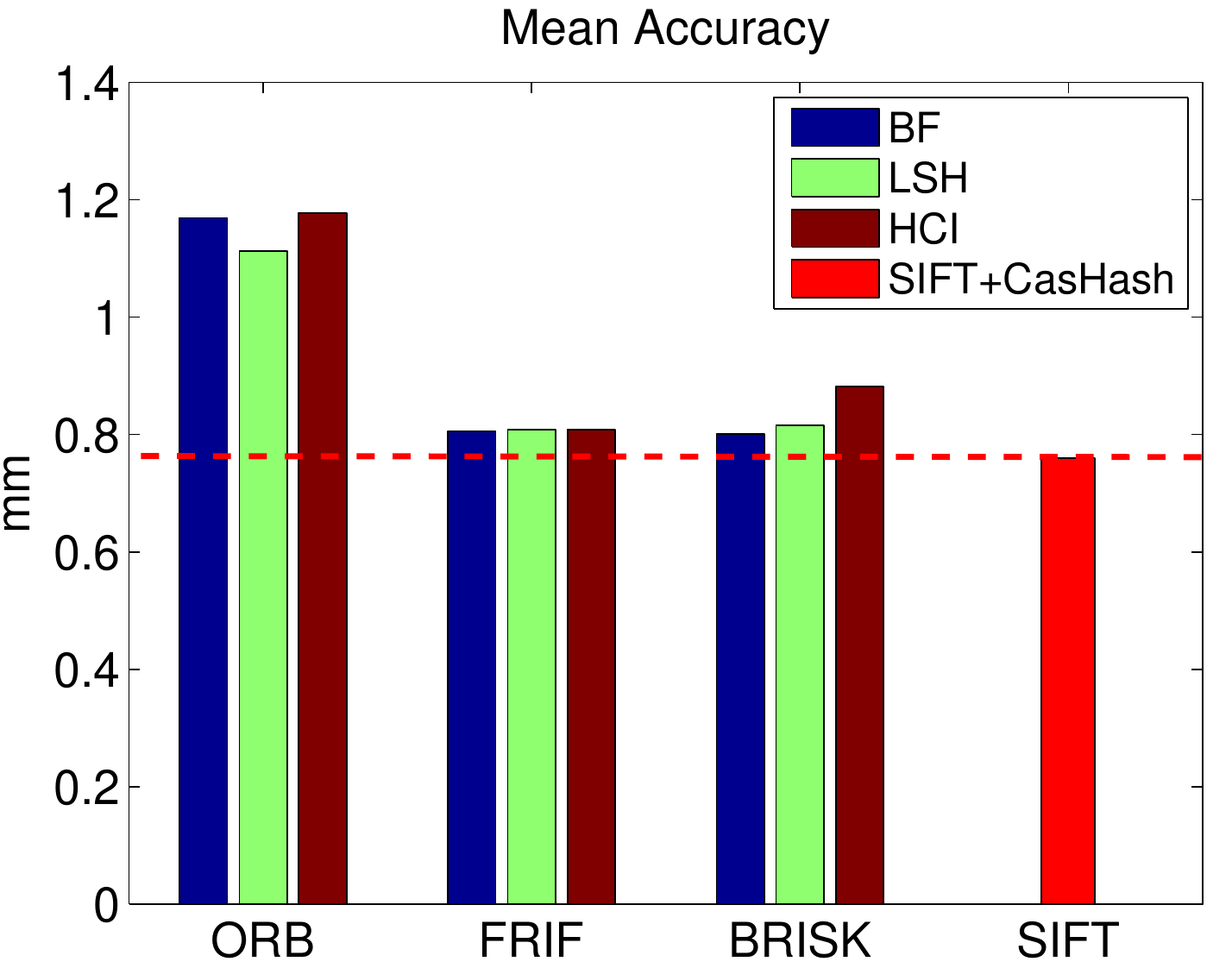}}
	\subfigure[]{
		\includegraphics[width=0.23\textwidth]{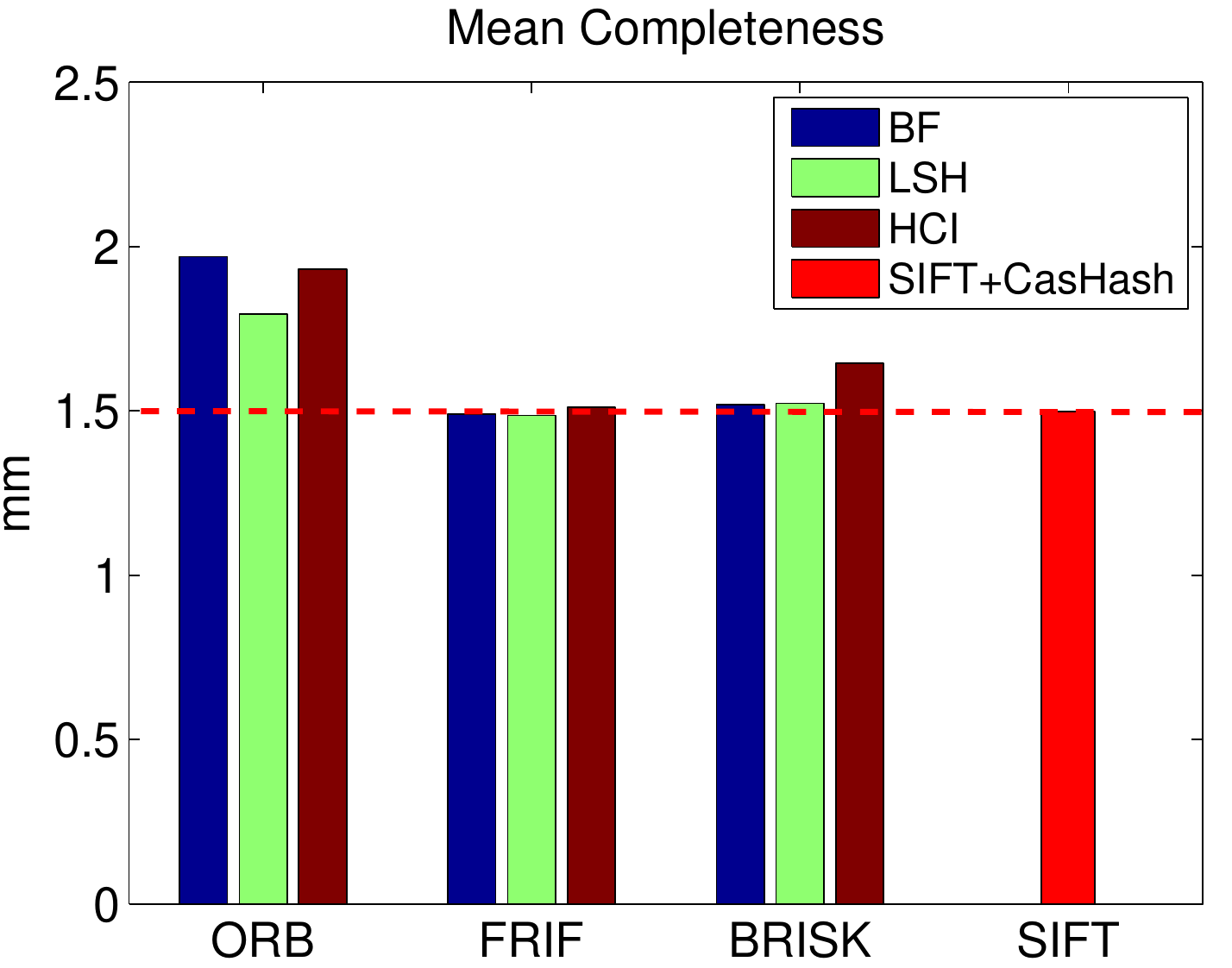}}
    \subfigure[]{
		\includegraphics[width=0.23\textwidth]{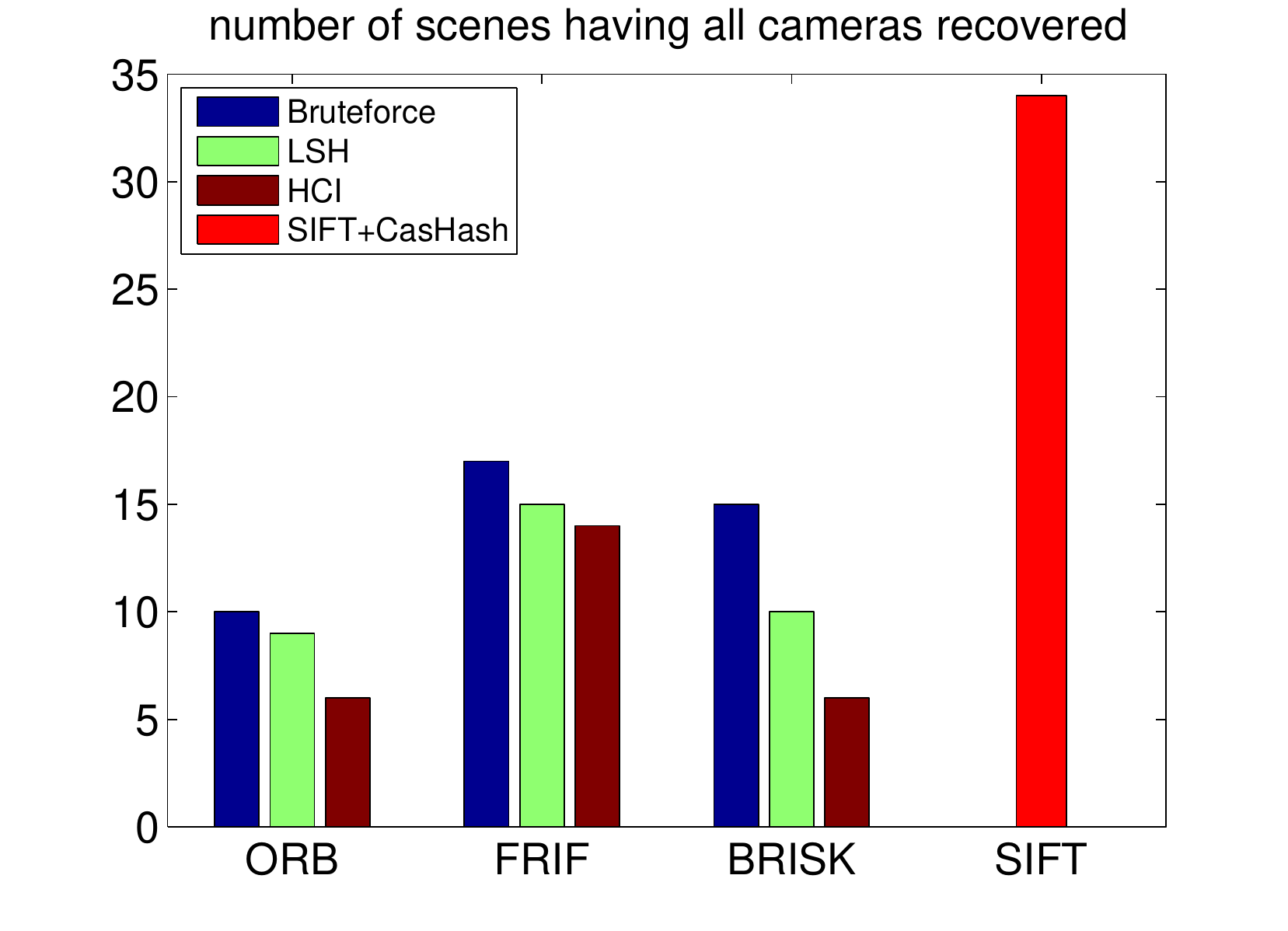}}
    \subfigure[]{
		\includegraphics[width=0.23\textwidth]{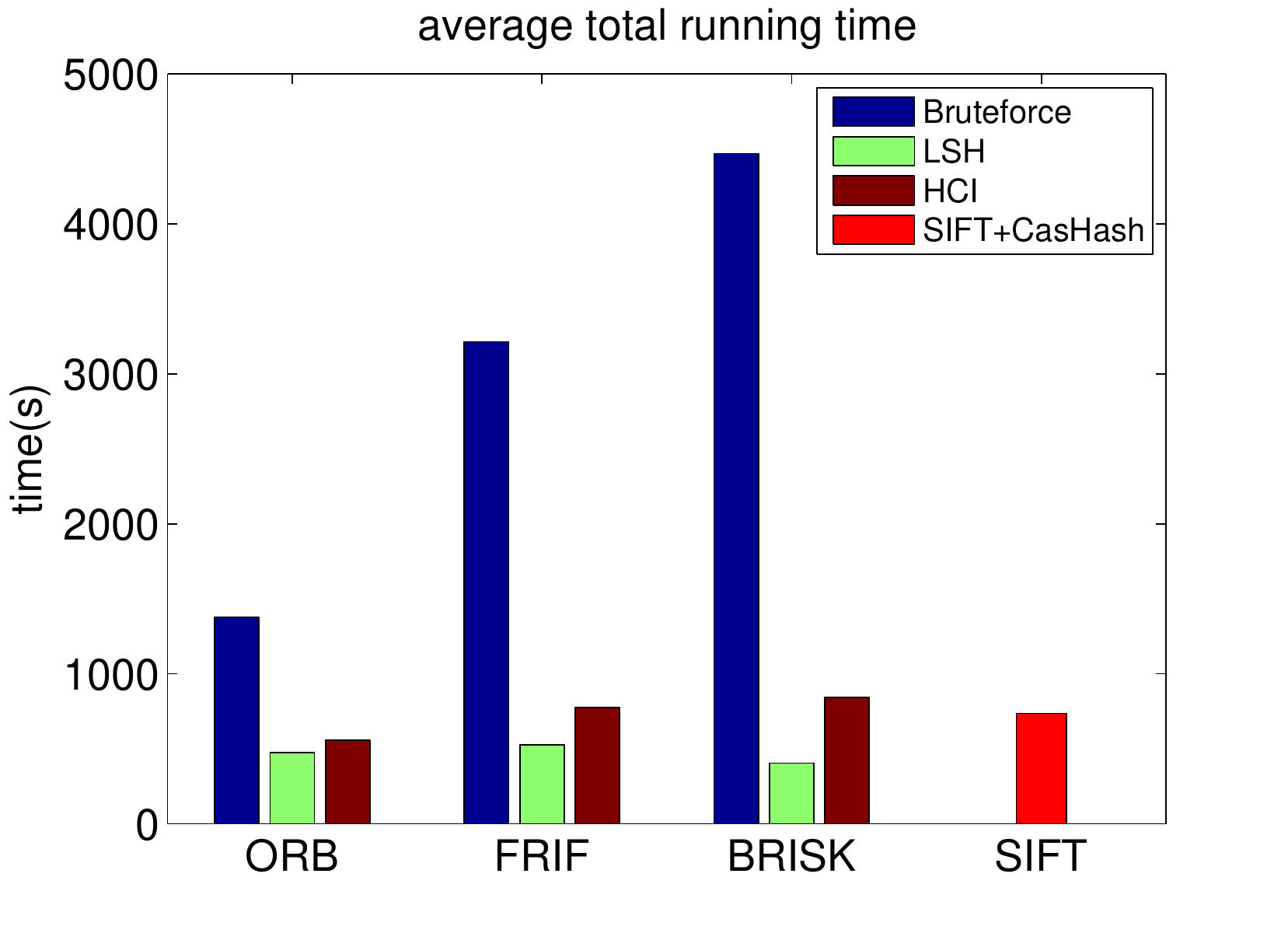}}
	\caption{A performance glance of different feature matching methods. (a) and (b) show the mean accuracy and mean completeness of 3D reconstruction on the DTU MVS dataset~\cite{Jensen_CVPR14}. The lower, the better. (c) depicts the number of scenes that all cameras have been successfully recovered by structure from motion based the feature matching results. (d) shows the average running times. \label{fig:introduction}}
\end{figure}

In these light weight tasks, feature matching is usually conducted on several hundreds of keypoints. In this case, a bruteforce, linear scan of nearest neighbors is efficient enough, thus favoring Hamming distance over Eculidean distance. However, in image based 3D reconstruction, high resolution images and large number of images are ubiquitous. It typically includes thousands or millions image matching operations~(quadratic in the number of input images), each of which encounters matching tens of thousands keypoints. For such a large scale feature matching problem, current solution is using SIFT and its approximate nearest neighbor~(ANN) searching method, for instance, KD-Tree~\cite{FLANN} and cascade hashing~\cite{CasHash}. Even with the most efficient Hamming distance, linear search of nearest neighbors is impractical according to our experimental results~( Fig.~\ref{fig:introduction}(d)). Apparently, ANN methods that are capable of dealing with binary descriptors are required too.

Most of such ANN methods are proposed recently~\cite{Trzcinski_PRL12,Muja_CRV12,Muja_PAMI14,Esmaeili_PAMI12}, and they are not well studied. This paper aims at a comparative study of the recently proposed binary features and ANN methods in the task of 3D reconstruction, which is a classical computer vision problem requiring lots of feature matching operations. It tries to answer the following questions: (1) Is binary feature still a premier choice for large scale feature matching problem? (2) If so, which one in the literature performs the best? (3) Which ANN method works the best? It is worth to point out that although there are many works on local feature evaluation in the literature, most of them are limited to the image matching level~\cite{Mikolajczyk_PAMI05,Aanaes_IJCV12,Miksik_ICPR12,Heinly_ECCV12}.

For this comparative study, a basic but typical 3D reconstruction system is implemented\footnote{We will make our system and evaluation code public available.}. By using this system, we can evaluate different binary features along with different feature matching methods. As a baseline feature matching method for comparison, SIFT matching with cascade hashing strategy~\cite{CasHash} is also evaluated. We choose to conduct evaluations on a recently proposed multiview stereo dataset~(DTU MVS)~\cite{Jensen_CVPR14}, which contains more than 100 different scenes with high resolution images captured from 49 or 64 viewpoints. Groundtruth 3D points are available, making it a good testbed for our purpose. We would like to first summarize our principal findings in the following and leave details in the remaining of this paper.
\begin{itemize}
\item[(1)] In terms of 3D reconstruction accuracy and completeness, Fig.~\ref{fig:introduction}(a)-(b) demonstrate that SIFT matching is better than matching of binary features. Among all the three tested binary features, FRIF~\cite{Wang_BMVC13} performs the best and only slightly worse than SIFT.
\item[(2)] As can be seen in Fig.~\ref{fig:introduction}(c), in terms of the number of recovered cameras by structure from motion, using SIFT matching is significant better than matching binary features. Using SIFT matching could recover all the cameras in 34 scenes, while using a binary feature, the best result is 17.
\item[(3)] As shown in Fig.~\ref{fig:introduction}(d), bruteforce matching of binary features is very time consuming. By using an ANN method, either float point or binary descriptor could achieve an acceptable time complexity. Among them, BRISK~\cite{Leutenegger_ICCV11} with LSH~\cite{Gionis_VLDB99,Lv_VLDB07} is the most efficient one, followed by ORB with LSH. Considering the tradeoff between accuracy and speed, FRIF with LSH is a good choice as it is only slightly worse than SIFT in performance, but more efficient.
\item[(4)] For ANN methods of binary features, Locality Sensitive Hashing(LSH) is consistently better than Hierarchical Clustering Index~(HCI)~\cite{Muja_CRV12}, both in terms of running time and accuracy, as shown in Fig.~\ref{fig:introduction}. Meanwhile, using LSH can obtain reconstruction results as good as using bruteforce matching.
\item[(5)] To some extent, 3D reconstruction does not rely too much on the number of recovered cameras. On the basis of the number of scenes whose cameras are all being recovered~(Fig.~\ref{fig:introduction}(c)), although FRIF is far below than SIFT, their 3D reconstruction performance does not differ too much~(Fig.~\ref{fig:introduction}(a) and Fig.~\ref{fig:introduction}(b)).
\end{itemize}

The remaining parts of this paper are organized as follows. In Section 2, we briefly describe our implemented 3D reconstruction system. Then, the evaluated local features and feature matching methods are introduced in Section 3 and Section 4 respectively. After describing the dataset and evaluation protocol in Section 5, we report and discuss results in Section~6. Section 7 concludes the paper.

\section{Pipeline of 3D Reconstruction}
To obtain the 3D points of an object or a scene by only using a number of images, the popular solutions~\cite{Agarwal_11,Frahm_10,Heinly_CVPR15} usually include three steps: feature matching across images, structure from motion~\cite{Snavely_SIGGRAPH06,MRF_SFM,ChangChangWu_SFM} and dense reconstruction~\cite{Furukawa_PAMI10}. Feature matching aims to find the so called feature tracks. In essential, a feature track corresponds to a 3D point, containing point correspondences across different images. For unordered and very large scale image collection, there is usually an additional preprocessing step, aiming to quickly find out possible overlapping image pairs so as to conduct feature matching only on these pairs to save matching time~\cite{Lou_ECCV12,Schonberger_CVPR15_1}. Structure from motion~(SFM) takes a number of feature tracks as input, and outputs a number of 3D points as well as some camera parameters of the input images. With the recovered cameras, dense reconstruction is applied to obtain a dense 3D point cloud as the reconstruction result. The system outputs include a number of 3D points of the scene and the estimated camera parameters of the input images. By comparing these outputs to the groundtruth, one can evaluate how good the system is, e.g., in terms of 3D reconstruction accuracy, completeness and successfully recovered cameras.

In this paper, we focus on the step of feature matching, studying its performance when using different binary features and the related matching methods. As a result, we fix the last two steps with typical methods: linear time incremental structure from motion~\cite{ChangChangWu_SFM} and PMVS~\cite{Furukawa_PAMI10} for dense reconstruction. Their source codes are provided and can be downloaded from their websites. Meanwhile, no preprocessing is used. In the following, we give a brief introduction to the evaluated features and their matching methods.

\section{Evaluated Local Features}
Since there are many binary descriptors as well as feature detectors in the literature, we try to avoid the combination problem of detector and descriptor by only choosing those methods containing both feature detector and descriptor. In the scope of binary descriptors, there are three such methods: ORB~\cite{Rublee_ICCV11}, BRISK~\cite{Leutenegger_ICCV11} and FRIF~\cite{Wang_BMVC13}. For float point descriptors, there are SIFT~\cite{LOWE_IJCV04}, SURF~\cite{Bay_CVIU08} and KAZE~\cite{KAZE_ECCV12}. As this paper is focused on studying the 3D reconstruction performance of binary descriptors, we only choose SIFT as the baseline and at the meantime choose all these three binary features for a comparative study.

\subsection{ORB}
To achieve scale and rotation invariance, ORB contains a multiscale FAST~\cite{FAST} detector and an intensity centriod based method for computing keypoint orientation. It implements a scale pyramid of the input image and detects FAST keypoints from all levels of the pyramid. A very loose threshold is firstly used to get as many FAST corners as possible. Then, the top $N$ corners with the highest Harris cornerness measure~\cite{Harris_1988} are kept, where $N$ is the expected number of keypoints. For each keypoint, a reference orientation is computed by taking the direction from the keypoint to the centriod of a local patch around the keypoint.

For feature description, ORB constructs a binary descriptor by intensity tests, similar to BRIEF~\cite{Calonder_PAMI11}. However, while BRIEF uses a random sampling pattern for intensity tests, ORB uses a learning based sampling pattern. In ORB, the intensity tests are selected from all possible candidates to contain as much information as possible while being less correlated to each other.

\subsection{BRISK}
The keypoint detector proposed in BRISK is based on AGAST~\cite{AGAST_ECCV10}, which is an effective extension of FAST detector. Instead of the Harris cornerness used in ORB, BRISK defines a FAST score as the saliency measure of a potential keypoint. Specifically, it is defined as the maximal threshold with which the point can be detected as a keypoint. Another difference lies in the implementation of scale space. BRISK uses two pyramids alternately, one for the octaves and the other for the intra-octaves, to cover a finer scale space than ORB does.

Given a sampling pattern with 60 sampling points regularly sampled from 4 concentric circles, BRISK divides their formed point pairs into long-distance pairs and short-distance ones. The long-distance pairs are used to compute an average local gradient to define the orientation of the keypoint, while the short-distance pairs are used for intensity tests to construct the binary descriptor. To deal with aliasing effects, the intensity of a sampling point is computed by filtering with a Gaussian kernel whose standard deviation is proportional to its distance to the keypoint, i.e., the central point of the sampling pattern.

\subsection{FRIF}
While both ORB and BRISK resort to FAST detector for efficient keypoint detection, FRIF was proposed to approximate the Laplacian of Gaussian~(LoG) with rectangular filters so that to compute its response very quickly. According to Mikolajczyk and Schmid's study~\cite{Mikolajczyk_01}, Laplacian is stable in characteristic scale selection and has been used in many feature detectors~\cite{Mikolajczyk_IJCV04,LOWE_IJCV04}. In FRIF, it approximates a LoG template by linear combination of four rectangles. Therefore, computing the LoG responses on pixels of an image just requires linear combination of four rectangular filtering results, which can be done efficiently based on integral images. To detect extrema of the approximated LoG responses across both spatial and scale spaces, FRIF implements an identical scale space as BRISK does and uses a similar strategy for non-maximum suppression as well as location refinement.

As far as the binary descriptor is concerned, FRIF uses a similar sampling pattern to BRISK, but proposes a mixed binary descriptor to achieve better performance. For each sampling point, it uses its neighboring points to conduct intensity tests to obtain a number of bits as part of the descriptor. It also uses some short-distance point pairs for intensity tests as the remaining part of the descriptor to capture complementary information. The long-distance point pairs are used to compute the keypoint orientation as in BRISK.

\subsection{SIFT}
SIFT constructs a Difference of Gaussian~(DoG) scale space to detect extrema across both spatial and scale spaces as keypoints. DoG scale space is constructed by subtracting neighboring images of a Gaussian scale space of the input image. The keypoint orientation is computed by accumulating a histogram of gradient orientations from a local circular region around the keypoint. The orientation corresponding to the largest bin in this histogram is taken as the keypoint orientation. Meanwhile, other orientations corresponding to the peak bins which are within 80\% of the largest one are also taken as the keypoint orientations.

For feature description, SIFT divides the scale and rotation normalized local patch around a keypoint into $4 \times 4$ grids. In each grid, it computes a histogram of gradient orientations with 8 bins. All these histograms are concatenated together and normalized to get a 128 dimensional float vector as the SIFT descriptor. To improve its robustness, the trilinear interpolation among spatial and orientation bins is utilized and a Gaussian weight is assigned to each pixel in the local patch.

\subsection{Implementation Details}
All the evaluated features have source codes available on the Internet. For ORB, we use the implementation supplied in the OpenCV 2.4.9. For BRISK and FRIF, we use the original implementations released by their authors respectively\footnote{BRISK: \href{http://www.asl.ethz.ch/people/lestefan/personal/BRISK}{http://www.asl.ethz.ch/people/lestefan/personal/BRISK}
	
~~~FRIF: \href{https://github.com/foelin/FRIF}{https://github.com/foelin/FRIF}}.
For SIFT, we use the implementation supplied in VLFeat~\cite{vlfeat}. For a fair comparison, we use the default parameters of SIFT~(since it is the baseline and has been stably used over 10 years) and tune the detector threshold for other binary features to make them have a similar average number of features. The reason for us to do so is that these binary features are proposed to address a relative small scale problem, thus using the default parameters recommended by their authors can only produce a very small number of features. This leads to a small number of matches that further degrades the performance of SFM and final 3D reconstruction results.

For feature matching, we search for the top two nearest neighbors for a query feature and use their distance ratio~(NNDR)~\cite{LOWE_IJCV04} to decide if two keypoints match or not. The threshold is set to 0.6, and mutual matching is imposed.

\section{Approximate Nearest Neighbor Search}
The bruteforce searching of nearest neighbors is only suitable for matching a small number of descriptors. However, in case of 3D reconstruction from multiple images, it usually involves thousands or even millions of image matching operations as it has to match any two of the input images. Therefore, it is necessary to use some scalable approximate nearest neighbor~(ANN) search methods for feature matching in this task. Traditionally, KD-Tree~\cite{Opt-KDTree} and hierarchical vocabulary tree~\cite{Nister_CVPR06} are used for float point descriptors. Cascade hashing~(CasHash)~\cite{CasHash} is a recently proposed one which we choose to use in our evaluation for its good performance. For binary descriptors, although computing their Hamming distance is fast, it is still time consuming when we have to deal with large scale matching problem. We will show this point in the experimental section~(cf. Fig.~\ref{fig:timing-result}). Thus ANN is still required for matching binary descriptors. Locality Sensitive Hashing~(LSH)~\cite{Gionis_VLDB99,Lv_VLDB07} and Hierarchical Clustering Index~(HCI)~\cite{Muja_CRV12} are two popular ones suitable for binary descriptors.

\subsection{Locality Sensitive Hashing}
The basic idea of Locality Sensitive Hashing~(LSH) is to use a set of locality sensitive hashing functions to map a float vector into a binary string, which is used as the address for a hash table to index the database. In this way, adjacent vectors in the original space are expected to be located in the same hash bucket~(corresponding to an address in the hash table) with a high probability. Therefore, ANN search of a given query feature just needs to index the corresponding hash bucket based on its LSH code and then rank the retrieved data by computing distances in the original data space. Since indexing the hash bucket can be executed in constant time and the retrieved data is only a few, such strategy is very efficient in finding approximate nearest neighbors. Although LSH is originally proposed for ANN search of float vectors, it can be naturally extended to the case of binary vectors. In this case, randomly selected elements in a binary vector are used to construct the hash table instead of the hashing functions used in the original LSH.

To increase the probability that the true nearest neighbors are within the retrieved data by indexing hash table~(i.e., lie in the same bucket as the query feature), multiple hash tables are used to produce a good candidate set. However, for high dimensional features, it usually requires too many hash tables to achieve a satisfactory performance. Lv et al.~\cite{Lv_VLDB07} proposed the multi-probe LSH to achieve the same performance with much less number of hash tables. They not only use the LSH code of a query feature to produce the candidate set, but also use the nearby hash buckets of its LSH code to improve the probability that candidate set contains the true nearest neighbors. Therefore, a common way is to use both multiple hash tables and multi-probe query strategy to achieve a higher performance with moderate memory cost.

\subsection{Hierarchical Clustering Index}
Hierarchical Clustering Index~(HCI) is a kind of data structure capable of fast matching of binary descriptors using Hamming distance. The idea is simple. It partitions and organizes the high dimensional binary data space into a hierarchical structure similar to the hierarchical vocabulary tree~\cite{Nister_CVPR06}. Specifically, given a dataset $\mathbb{D}$, it first randomly selects $m$ points in $\mathbb{D}$ and clusters all points in $\mathbb{D}$ by assigning them to the nearest selected point in Hamming space, where each such cluster is called a branch. This procedure is repeated iteratively for each branch until the size of cluster is smaller than a predefined threshold~(leaf size). After building hierarchical clustering tree for the given dataset, for a query binary feature, it first conducts tree traversal to get a small candidate set and then returns the nearest neighbors according to the Hamming distances between the query feature and elements in the candidate set. Similar to KD-Tree, randomly building multiple such hierarchical trees can largely boost its search performance.

\subsection{Cascade Hashing}
For our baseline SIFT descriptor, we use the recently proposed CasHash~\cite{CasHash} for ANN searching. It was reported with better performance than the previous widely used KD-Tree in the task of 3D reconstruction.

Essentially, CasHash uses two steps of LSH to accelerate the process of ANN searching. In the first step, it uses short LSH code~($m$ bits) along with multi-table strategy~($L$ hash tables) to quickly eliminate a large proportion of non-matches, returning a relatively small amount of potential matches in a constant time. Since the number of remaining candidates is still too large to be effectively searched, it utilizes a second step of LSH with longer code~($n$ bits). In this way, both the query feature and the candidate features returned in the first step are mapped into a $n$ bits Hamming space. Then, it builds a hash table to index the candidate features by setting their Hamming distances to the query feature as hash key values. Based on this hash table, it can efficiently return the top $k$ nearest neighbors of the query feature in this $n$ dimensional Hamming space. Finally, the nearest neighbors of the query feature are obtained by re-ranking them according to Eculidean distances in the original feature space.

\subsection{Implementation Details}
For LSH and HCI, we use their implementations in the FLANN library~\cite{FLANN} in our experiments. According to the results on several scenes that we randomly checked, the default parameter settings of LSH and HCI do not perform well in our task. As a result, we set them based on these randomly selected scenes. Due to the large size of the evaluated dataset, we can not check all the testing scenes and tune the parameters accordingly. For LSH, we use the multi-table, multi-probe LSH, and set the number of hash tables as 4, the multi-probe level as 1, the LSH code length as 24. For HCI, we use 2 hierarchical clustering trees, each of which has 48 branches in each level and has a leaf size of 150.

For the CasHash, we use the source code supplied by its authors\footnote{\href{http://www.nlpr.ia.ac.cn/jcheng/papers/CasHashing.tar.gz}{http://www.nlpr.ia.ac.cn/jcheng/papers/CasHashing.tar.gz}} along with its default parameters. There are $L = 6$ hash tables with $m = 8$ bits LSH code in the first step, and $n = 128, k = 6$ in the second step.

\section{Experimental Setup}
\textbf{Dataset}: We choose to evaluate the 3D reconstruction performance of different binary features on a recently published multiview stereo dataset, known as the DTU MVS dataset~\cite{Jensen_CVPR14}. It contains a total number of 124 different scenes, covering a wide range of objects and surface materials. For each scene, it collects images of $1600 \times 1200$ resolution from 49 or 64 different viewpoints, with 8 different illumination conditions. Among these scenes, 80 scenes contain necessary information~(i.e., observability mask) that is required for the evaluation of reconstruction results as Jensen et al. did~\cite{Jensen_CVPR14}. In this paper, we use the scenes with 49 views, which occupy 58 out of all 80 scenes. We do not study effects of different lighting conditions, so we just use the subset with all lights on.

Due to the fact that our evaluated system is fully automatic and uses the self-calibration to decide the camera parameters, the coordinate system of the reconstructed 3D points can be any of those recovered cameras. The reconstructed coordinate system and the supplied reference coordinate system are related by a 3D similarity transformation~(scaling, rotation and translation). Therefore, we have to firstly register the reconstructed 3D points to the reference scans~(groundtruth) obtained by a structure light scanner which are supplied in the dataset. To this end, we manually selected three corresponding 3D points between the reconstructed one and the groundtruth. Then, they are used to estimate a similarity transformation to register the reconstructed 3D points.

\textbf{Evaluation Protocol}: After registering the reconstructed 3D points to the reference coordinate system, we use the supplied code in the dataset for performance evaluation. The evaluation protocol is based on that of~\cite{Seitz_CVPR06}, with some modifications to make it unbiased and better at handling missing data and outliers. Basically, it adopts an observability mask so that the evaluation is only focused on the visible part of the scene. Please refer to~\cite{Jensen_CVPR14} for more details about how to obtain such masks.

As in~\cite{Seitz_CVPR06,Jensen_CVPR14}, accuracy and completeness are used as quality measures of a reconstruction. According to their definitions, given a reconstruction and the structured light reference, the accuracy is computed as the distance from the reconstruction to the reference scan. On the contrary, the completeness is computed as the distance from the reference scan to the reconstruction. For each 3D point in one~(either the reconstructed 3D points or the reference 3D points), its distance to the other is computed as the closest distance to all the 3D points in the other.

There are two situations that are commonly encountered in 3D reconstruction which could induce bias if they are not treated properly. One is that there are usually more 3D points in the textured regions, while the other one is outliers. We use the same strategy as in~\cite{Jensen_CVPR14} to deal with these problems. The first issue is addressed by subsampling, i.e., the reconstructed 3D points are subsampled so that any two points have a distance larger than 0.2mm. For the second issue, those points with large errors which could be outliers are simply removed. Specifically, the points whose distances are larger than 20mm are removed when computing accuracy and completeness. The mean accuracy and completeness are recorded to evaluate the quality of a reconstruction. The evaluation code implemented with these considerations and the dataset can be downloaded on: \href{http://roboimagedata.compute.dtu.dk}{http://roboimagedata.compute.dtu.dk}

All experiments reported in this paper are conducted in a laptop with Intel 2.5GHz CPU and 8GB memory.

\section{Results and Analysis}

\subsection{Performance}
In our implemented 3D reconstruction system, after feature matching, the linear time incremental SFM~\cite{ChangChangWu_SFM} is firstly conducted to recover the parameters of cameras which are further input to PMVS~\cite{Furukawa_PAMI10} to obtain 3D points of the scene. Therefore, besides reporting performance on the final 3D points, we also report the performance of SFM.

\begin{figure}[htb]
	\centering \subfigure[]{
		\includegraphics[width=0.23\textwidth]{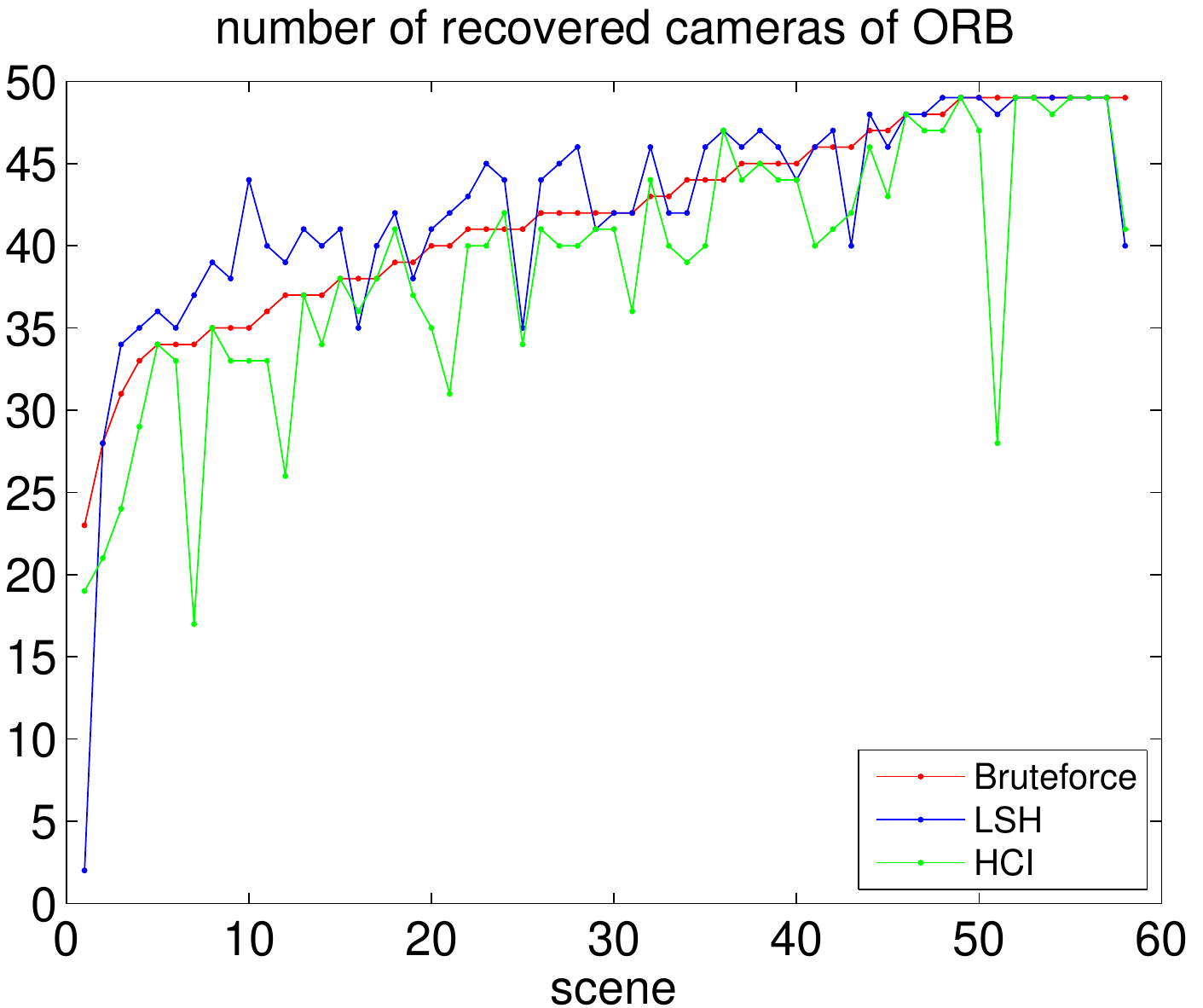}}
	\subfigure[]{
		\includegraphics[width=0.23\textwidth]{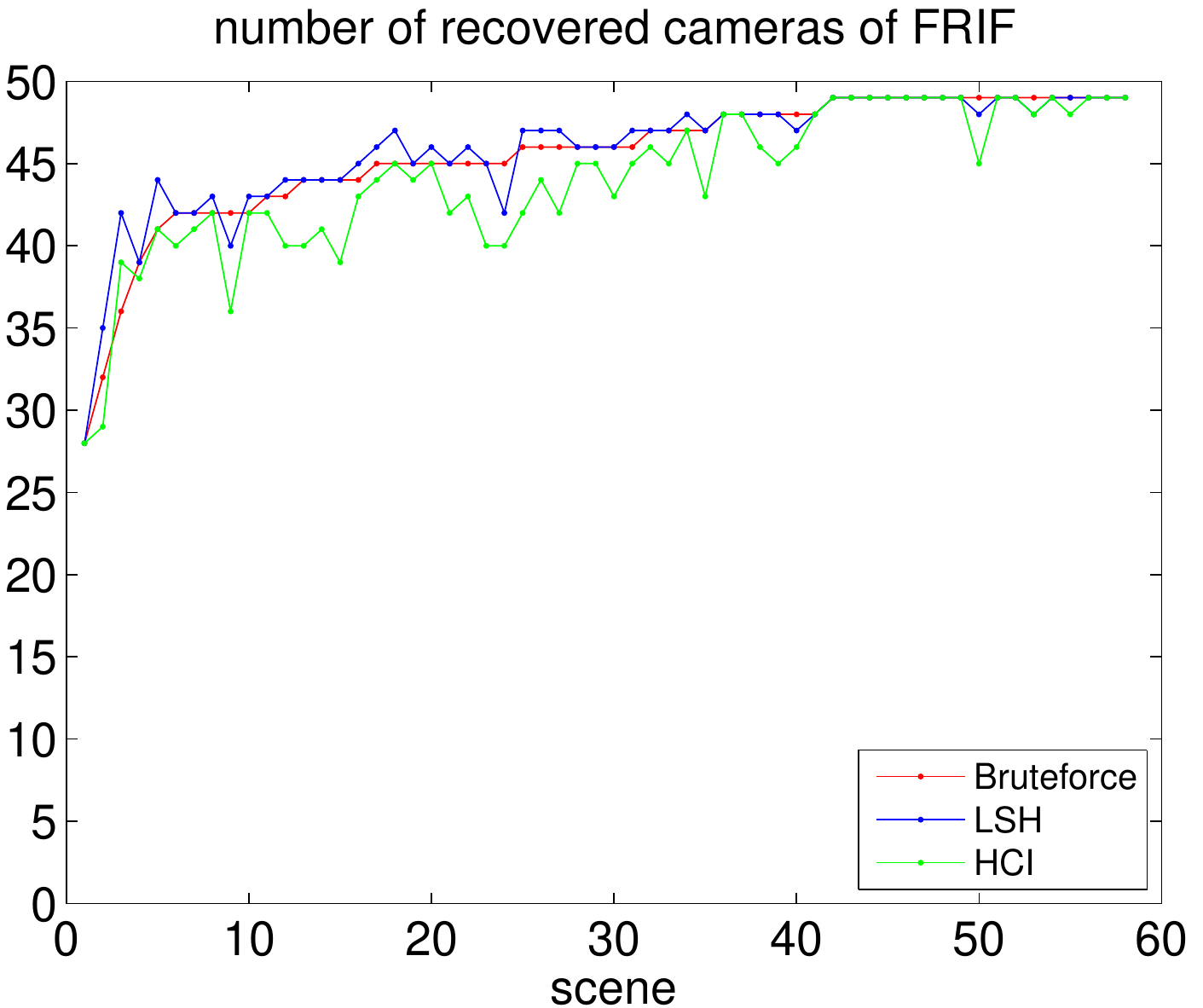}}
	\subfigure[]{
		\includegraphics[width=0.23\textwidth]{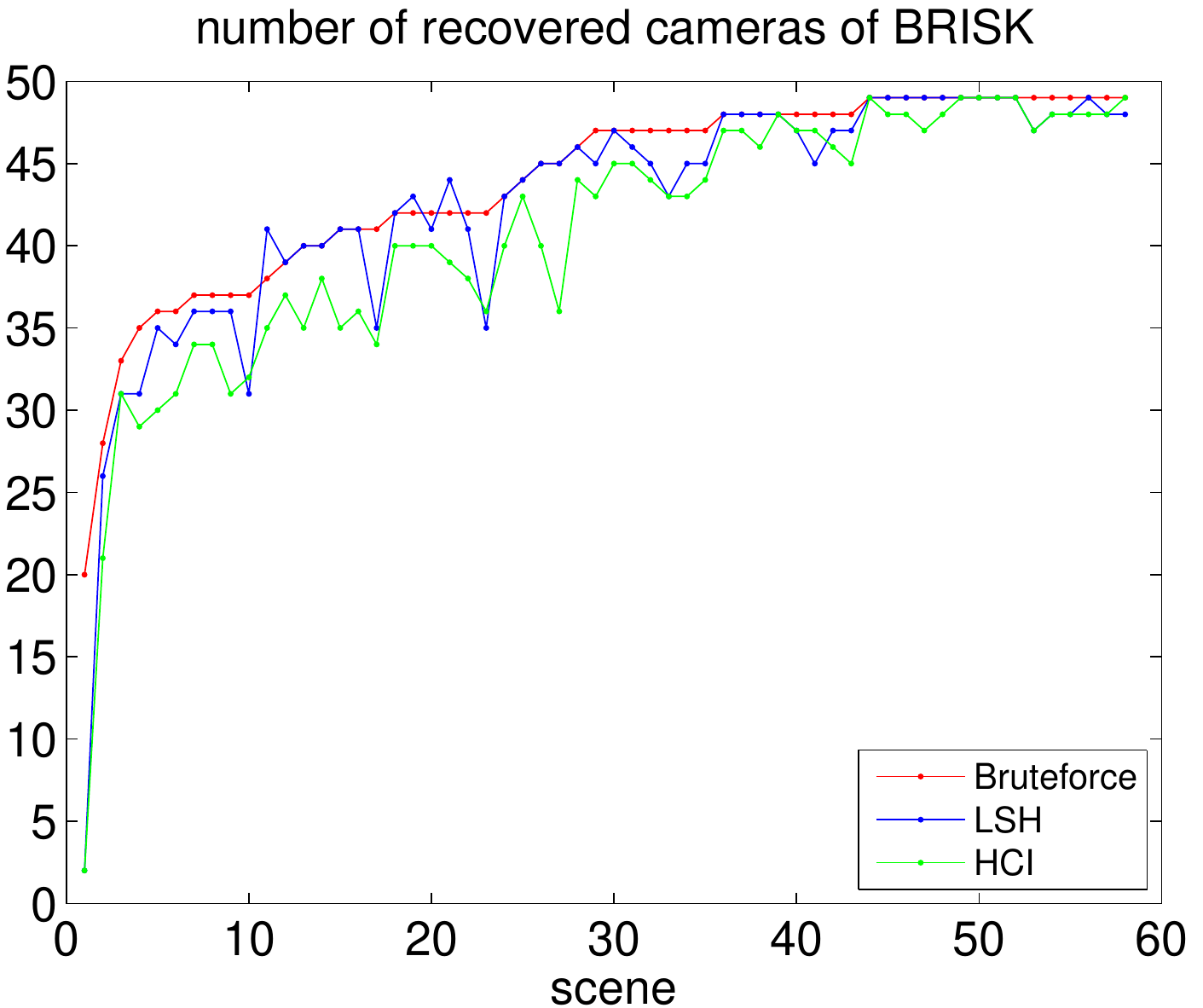}}
	\subfigure[]{
		\includegraphics[width=0.23\textwidth]{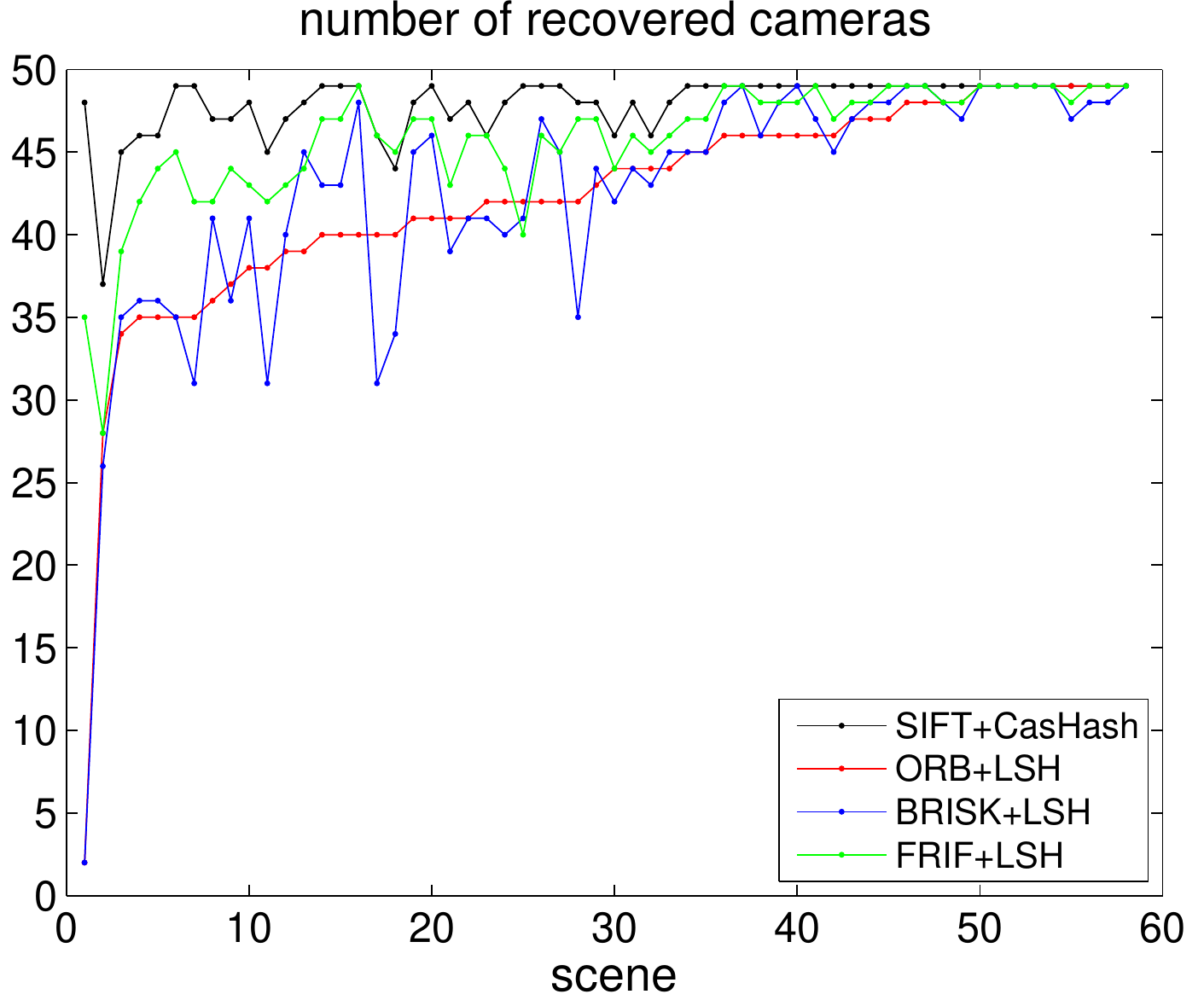}}
	\caption{Number of recovered cameras by structure from motion. (a)-(c) are the results of different matching methods for ORB, FRIF and BRISK respectively. For a clear comparison, (d) draws the results across different local features, where the results of LSH are shown with the binary features as it is the best according to (a)-(c). In order to get a better visual illustration, the tested scenes are rearranged in each subfigure so that the result is non-descending for bruteforce in (a)-(c) and for ORB+LSH in (d).  \label{fig:cam-result}}
\end{figure}

Fig.~\ref{fig:cam-result} shows the number of successfully recovered cameras by SFM. In Fig.~\ref{fig:cam-result}(a)-(c), they draw three curves for ORB, FRIF and BRISK respectively. Each of these curves corresponds to a specific feature matching method, including LSH, HCI and bruteforce as the baseline comparison. For a clearer visual illustration, the abscissas of these subfigures are rearranged so that the number of recovered cameras is non-descending for the bruteforce method. From them, we can find that LSH is consistently better than HCI independent of the used binary features. When using ORB and FRIF, LSH is even better than bruteforce as it results in more cameras being recovered by SFM. While for BRISK, LSH is a little worse than bruteforce. When only considering the number of fully recovered scenes~(i.e., the scene has all cameras being recovered), bruteforce is better than LSH for all tested binary features, which can also be read from Fig.~\ref{fig:introduction}(c). In Fig.~\ref{fig:cam-result}(d), it compares the performance of different local features. Due to the good performance of LSH according to Fig.~\ref{fig:cam-result}(a)-(c), it is chose to shown in Fig.~\ref{fig:cam-result}(d) with different binary features to make a comparison to SIFT matching with CasHash. Its abscissa is also rearranged so that the number of recovered cameras is non-descending for ORB. Overall, SIFT gets a significantly better result than all the binary features. Out of the 58 tested scenes, it recovers more than 44 cameras for 57 scenes, the remaining one scene has 37 cameras being recovered. What is more, it successfully recovers all cameras for 34 scenes, which is much better than other binary features~(also shown in Fig.~\ref{fig:introduction}(c)).

\begin{figure*}[htb]
	\centering
	\subfigure[]{
	\begin{minipage}{0.24\textwidth}
		\centering
		\includegraphics[width=1\textwidth]{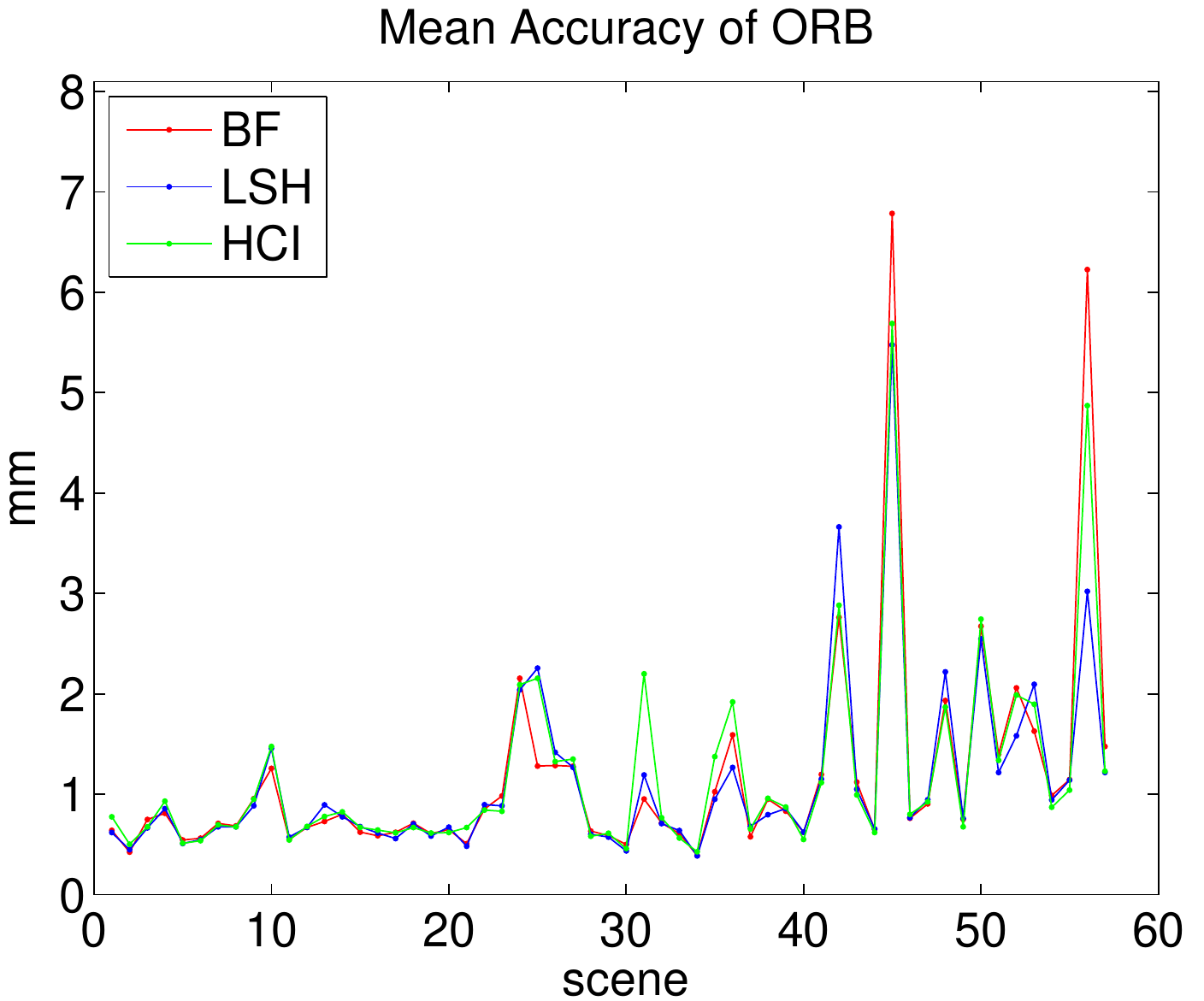}
		\includegraphics[width=1\textwidth]{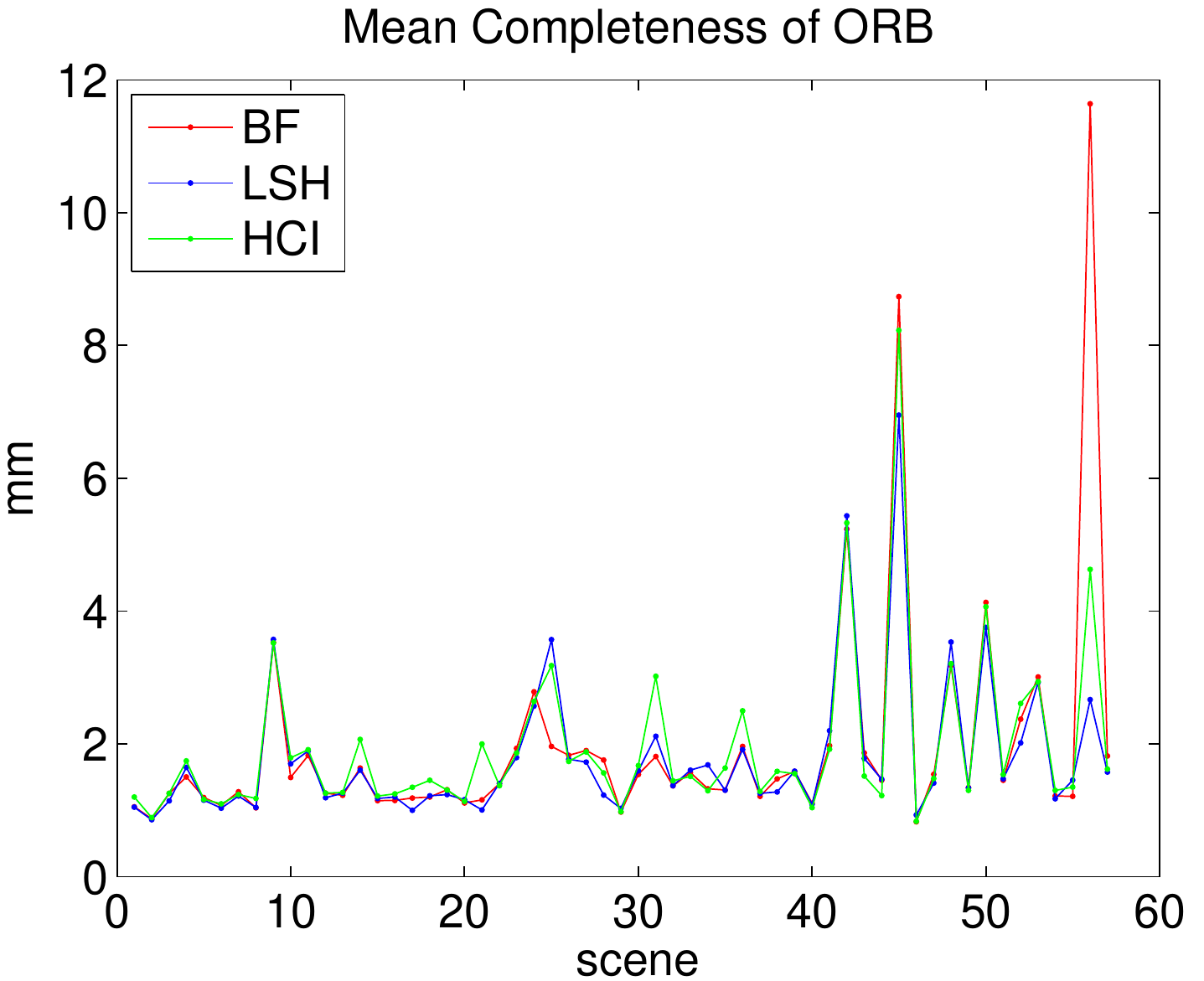}
	\end{minipage}}
	\subfigure[]{
	\begin{minipage}{0.24\textwidth}
		\centering
		\includegraphics[width=1\textwidth]{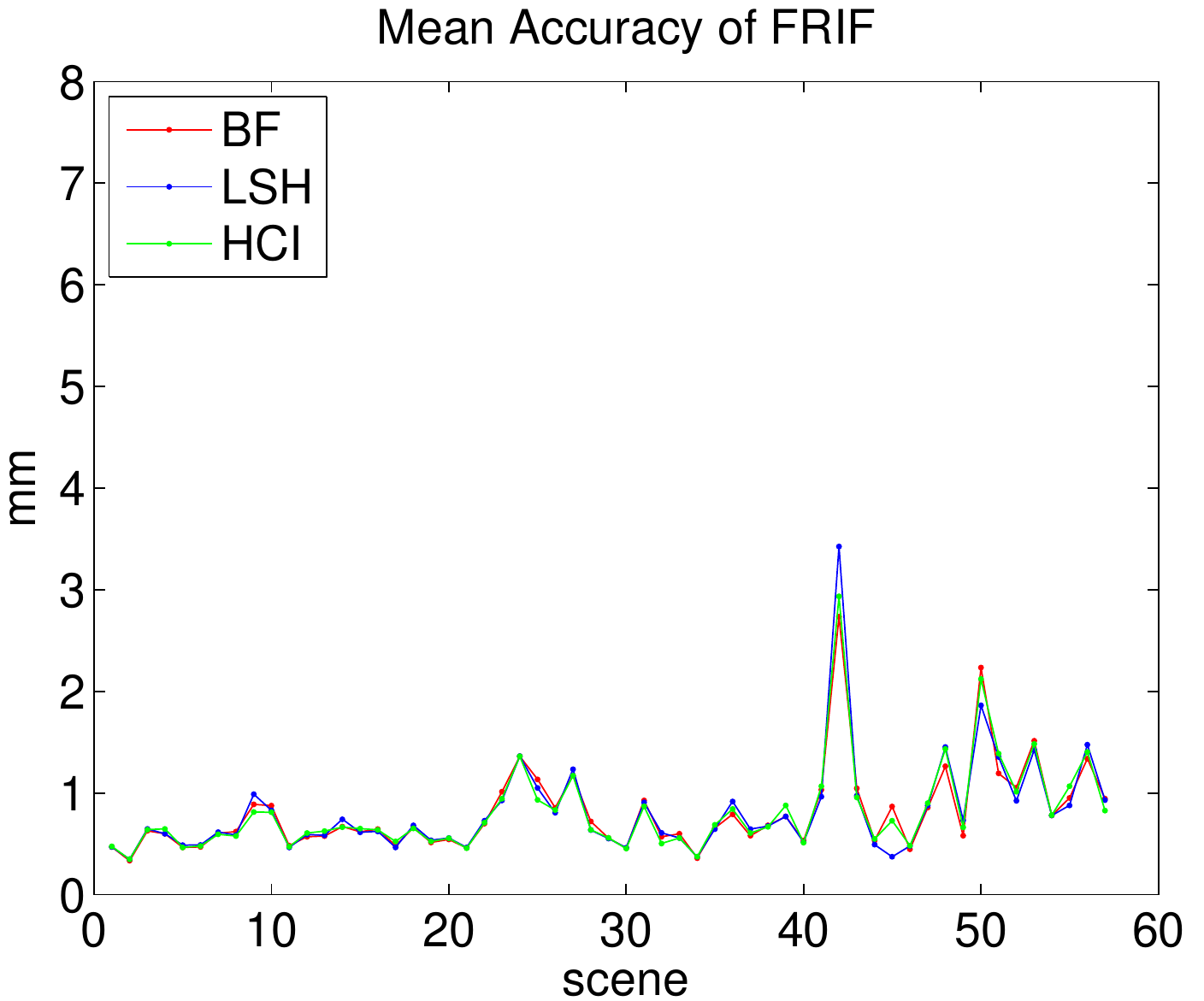}
		\includegraphics[width=1\textwidth]{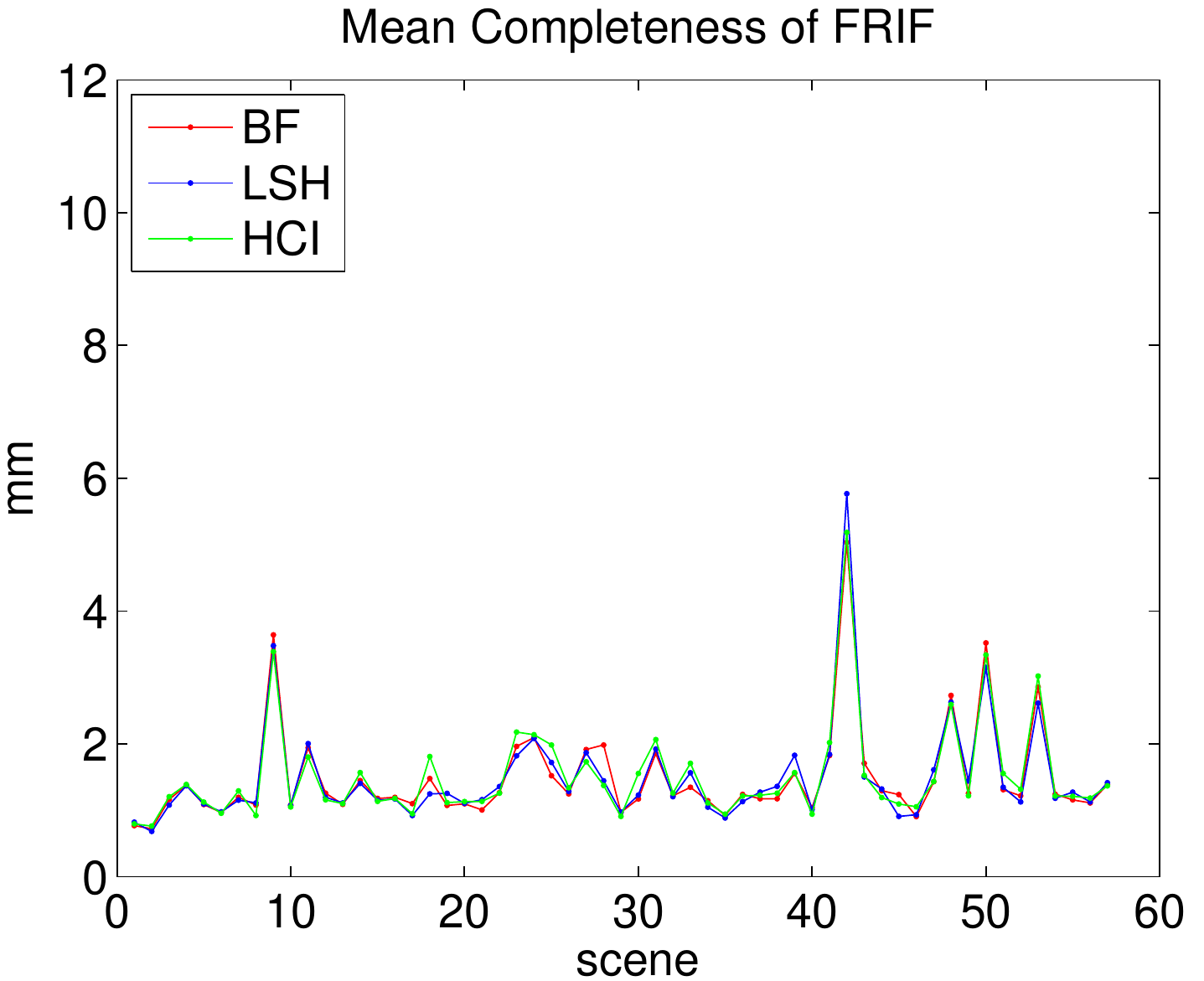}
	\end{minipage}}
	\subfigure[]{
	\begin{minipage}{0.24\textwidth}
		\centering
		\includegraphics[width=1\textwidth]{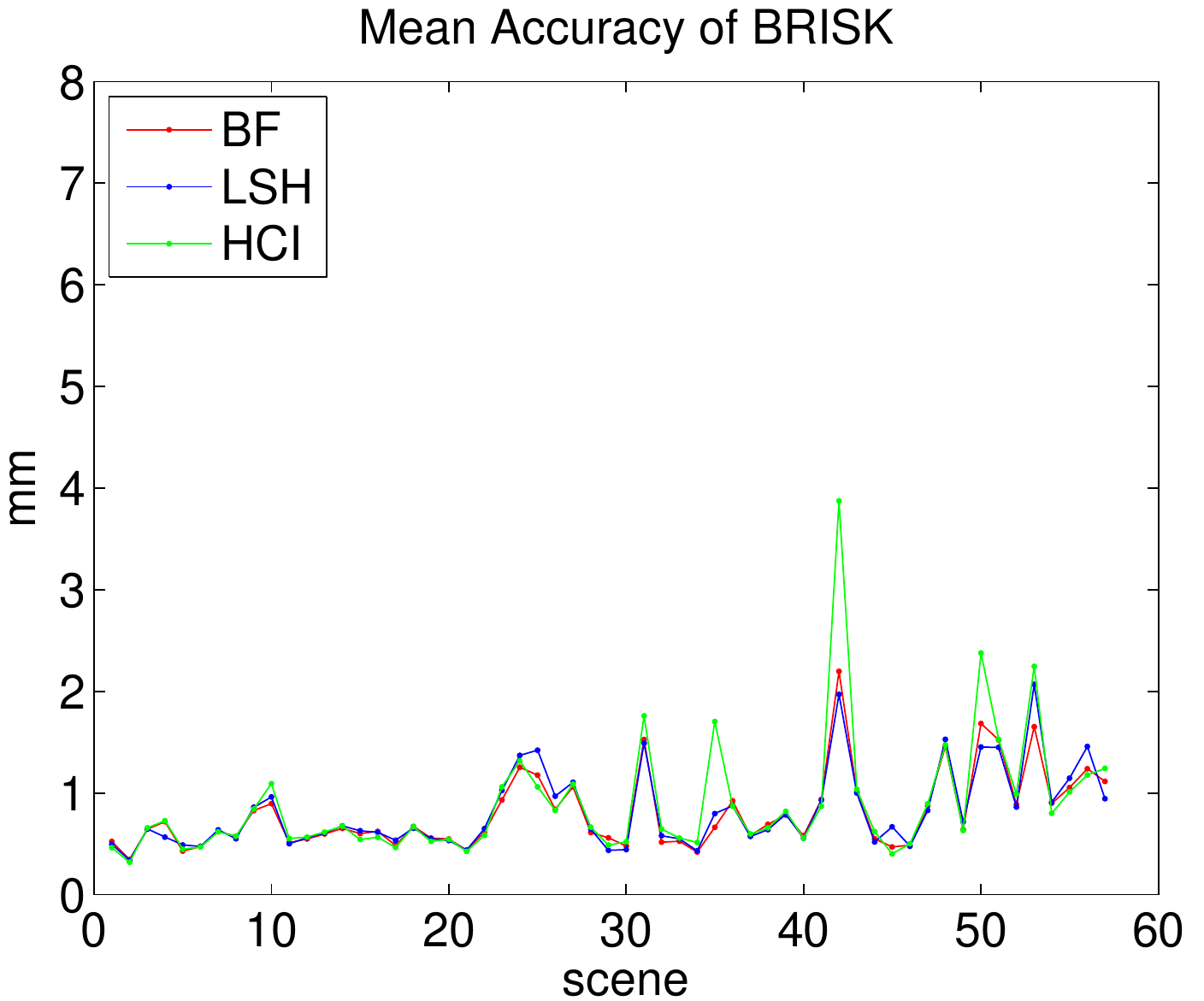}
		\includegraphics[width=1\textwidth]{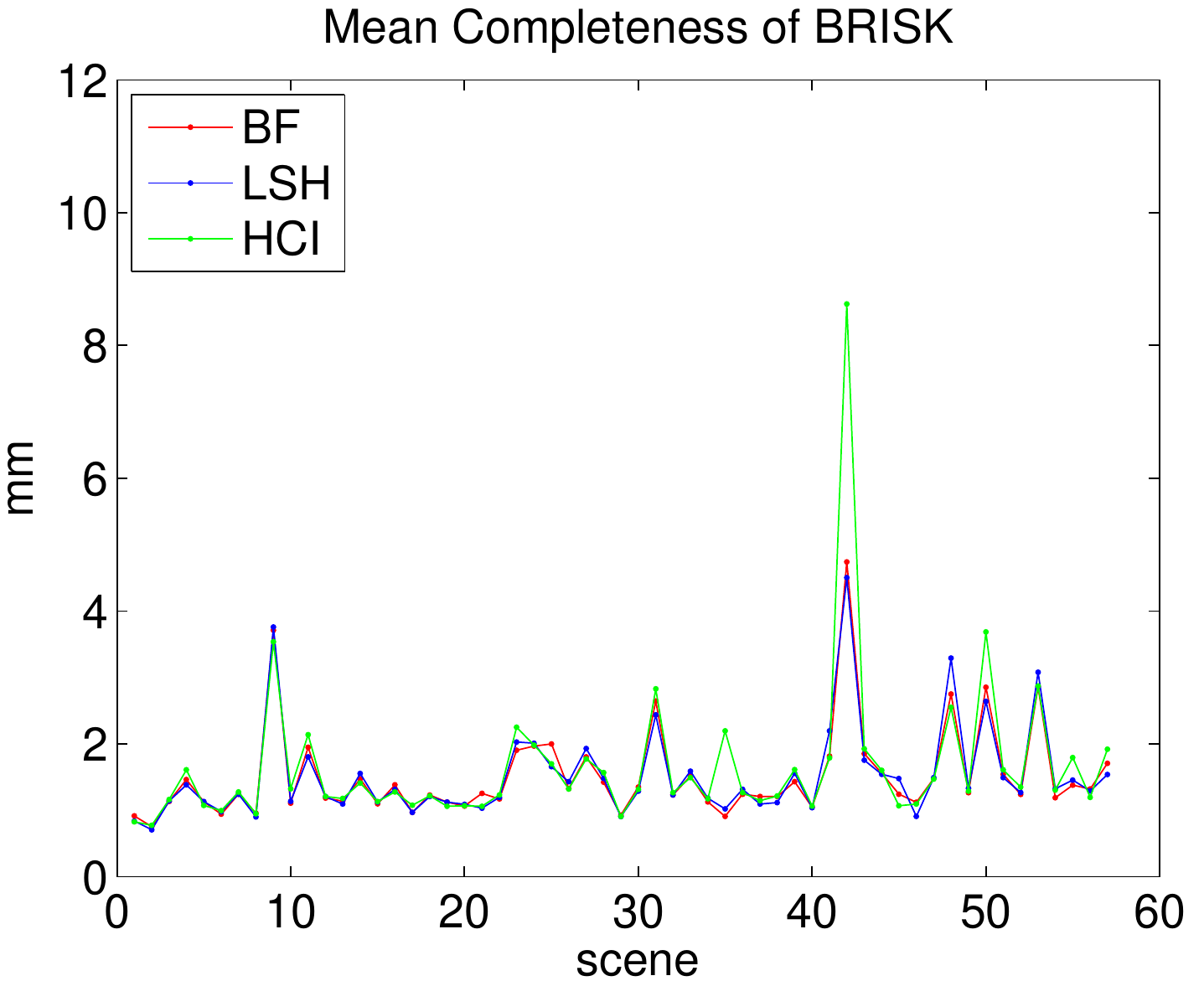}
	\end{minipage}}
	\subfigure[]{
	\begin{minipage}{0.24\textwidth}
		\centering
		\includegraphics[width=1\textwidth]{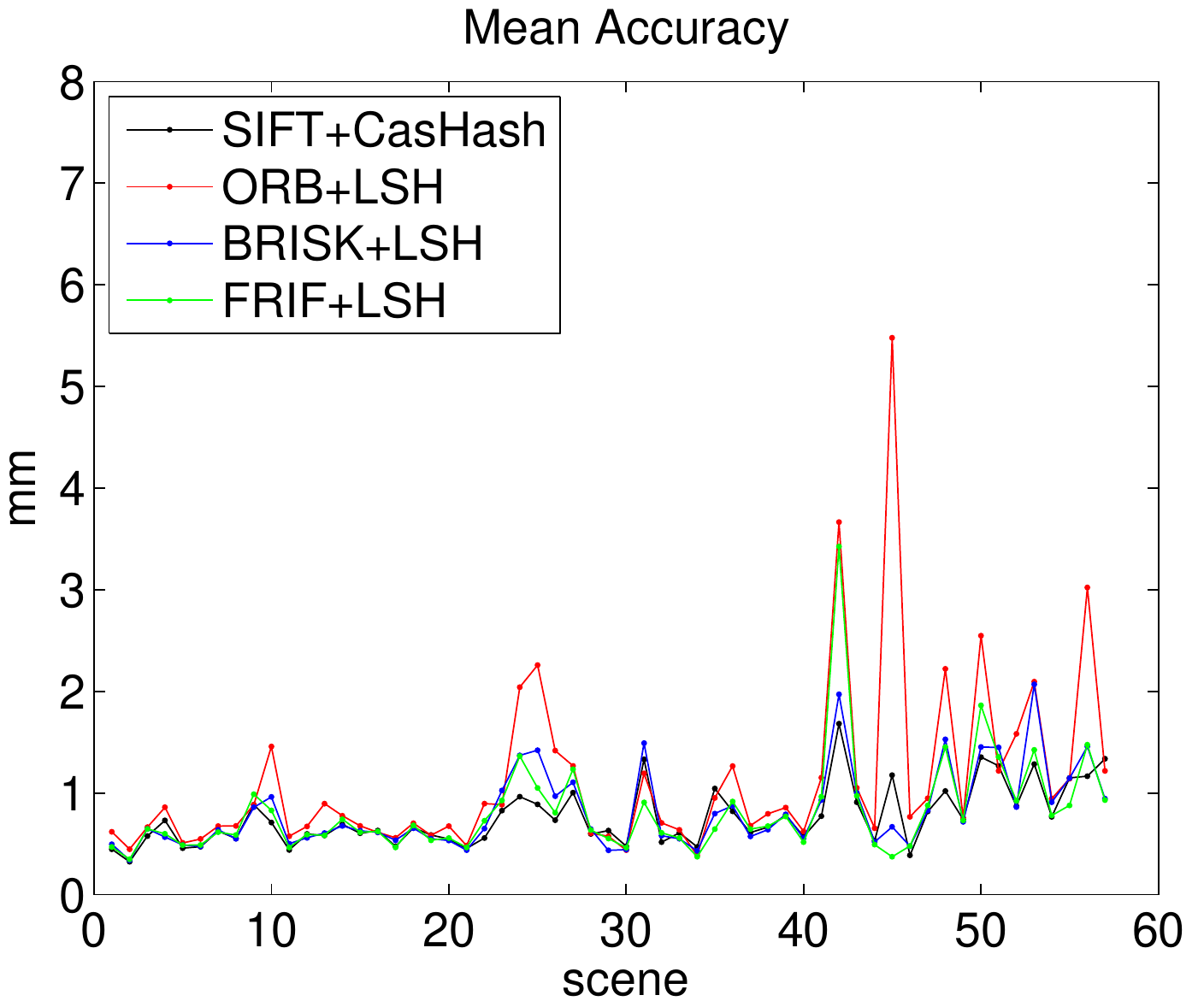}
		\includegraphics[width=1\textwidth]{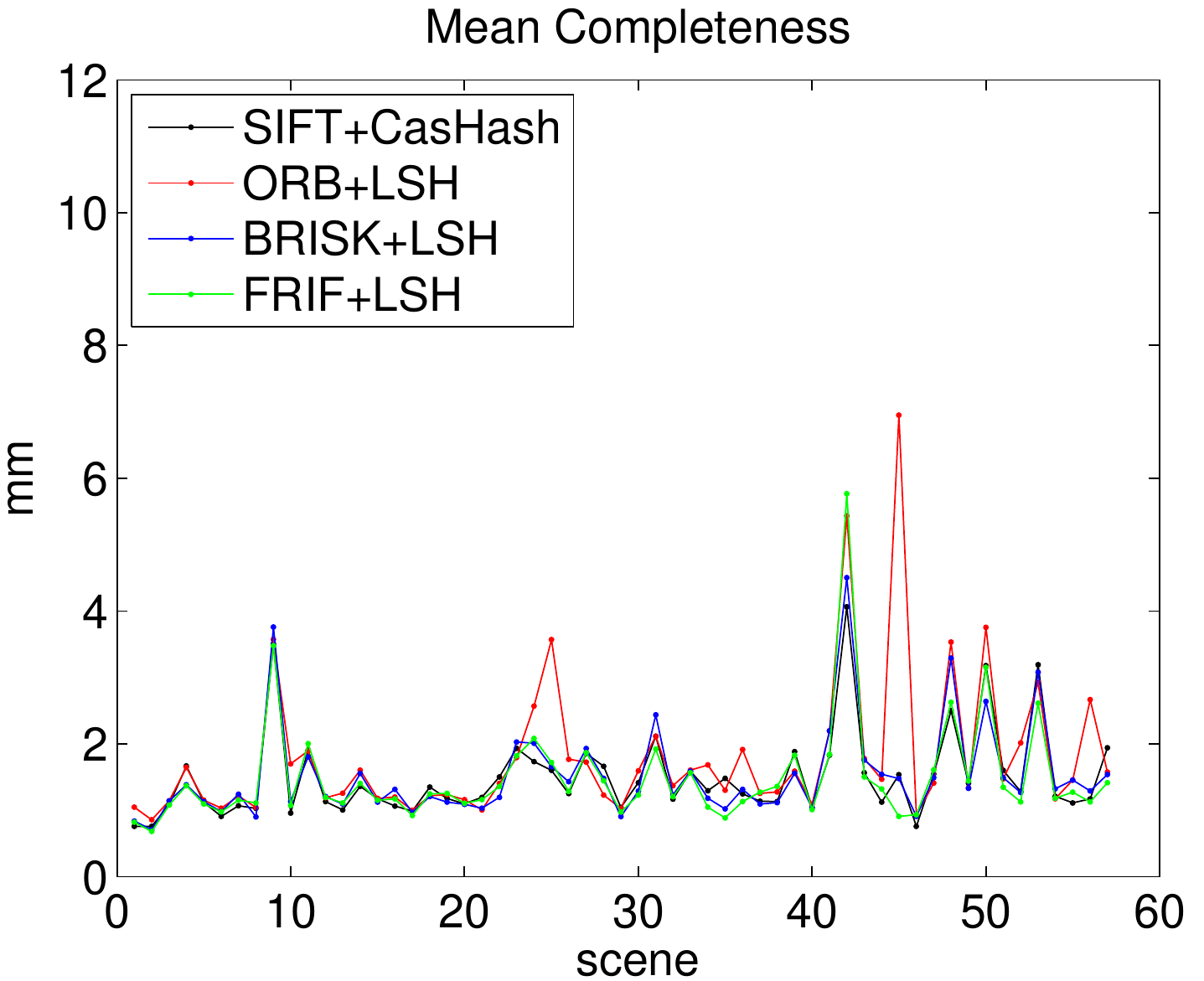}
	\end{minipage}}
	\caption{3D reconstruction performance for different local features and their matching methods. The top row shows the mean accuracies for different scenes, while the bottom row is the mean completeness. (a)-(c) are the results of ORB, FRIF and BRISK respectively by using different matching methods. (d) gives the comparison results across all the evaluated local features, where LSH is utilized for binary features due to its good performance. \label{fig:accu_and_complete}}
\end{figure*}

Fig.~\ref{fig:accu_and_complete} shows the mean accuracy and completeness for the tested scenes. For a specific curve, each point corresponds to the result of a scene. There is one scene that 3D reconstruction would fail when using some of the evaluated feature matching methods, thus, it is excluded in Fig.~\ref{fig:accu_and_complete}. In other words, Fig.~\ref{fig:accu_and_complete} actually gives the results of 57 tested scenes, for each of which all the evaluated methods could lead to a 3D point cloud of the scene. The results of ORB, FRIF and BRISK are shown in Fig.~\ref{fig:accu_and_complete}(a)-(c) respectively. Top row shows the mean accuracy of each tested scene obtained by different feature matching methods, while the bottom row is the mean completeness. Obviously, HCI performs the worst. LSH has a similar performance to bruteforce. As far as the binary feature is concerned, using ORB usually leads to worse results than using FRIF or BRISK. The latter two have a similar performance, with FRIF slightly better. To have a clearer comparison among different local features, we plot together the results of all the evaluated features in Fig.~\ref{fig:accu_and_complete}(d), where binary features use LSH. It is apparent that ORB is the worst, followed by BRISK. SIFT performs the best, and closely followed by FRIF. This can also be read from the average results shown in Fig.~\ref{fig:introduction}(a) and Fig.~\ref{fig:introduction}(b).

Combining the results in Fig.~\ref{fig:cam-result} and Fig.~\ref{fig:accu_and_complete}, it is interesting to see that although FRIF does not perform as well as SIFT in SFM, their final results in 3D reconstruction are similar. By inspecting Fig.~\ref{fig:cam-result}(d), we can find that FRIF almost has an identical trend as SIFT in terms of the number of recovered cameras. It just recovers several less cameras than that recovered by SIFT. Considering the fact that FRIF has a similar performance in 3D reconstruction as SIFT, this indicates that some recovered cameras are redundant for PMVS.  This explanation is reasonable, because there are overlaps between the viewpoints of different cameras. Therefore, in some cases, missing information caused by one unrecovered camera can be compensated by its neighboring cameras that have been successfully recovered. Consequently, it is not always necessary to recover all cameras to reconstruct the whole scene. However, a common sense is that the more recovered cameras, the better for 3D reconstruction. At least, it will not degrade the performance if it does not help as shown by our results. To sum up, recovering as many cameras as possible is a sufficient but not necessary condition for a good 3D reconstruction. There are some key camera positions with respect to the imaged scene.

Besides the higher discriminative ability of descriptor, another reason for the superior performance of FRIF over BRISK and ORB may be lie in its keypoint detector. FRIF is to detect blob like keypoints as SIFT does, while both BRISK and ORB detect keypoints based on FAST detector that responses largely on corners. Due to the fact that both FRIF and SIFT are consistently better than BRISK and ORB, we conclude that blob like keypoints could be more suitable for tasks of SFM and 3D reconstruction.

\subsection{Timing}
\begin{figure}[t]
	\centering \subfigure[]{
		\includegraphics[width=0.23\textwidth]{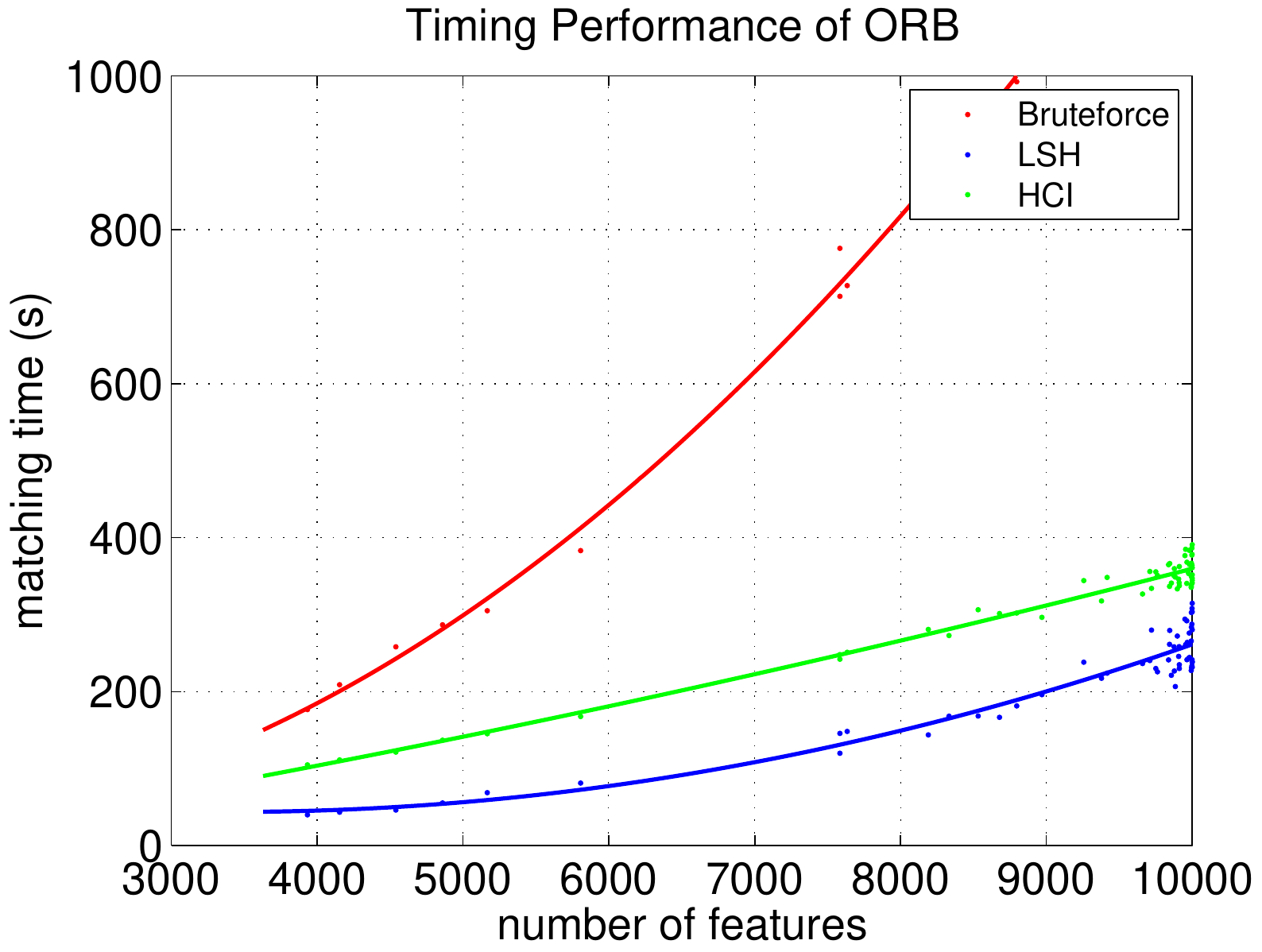}}
	\subfigure[]{
		\includegraphics[width=0.23\textwidth]{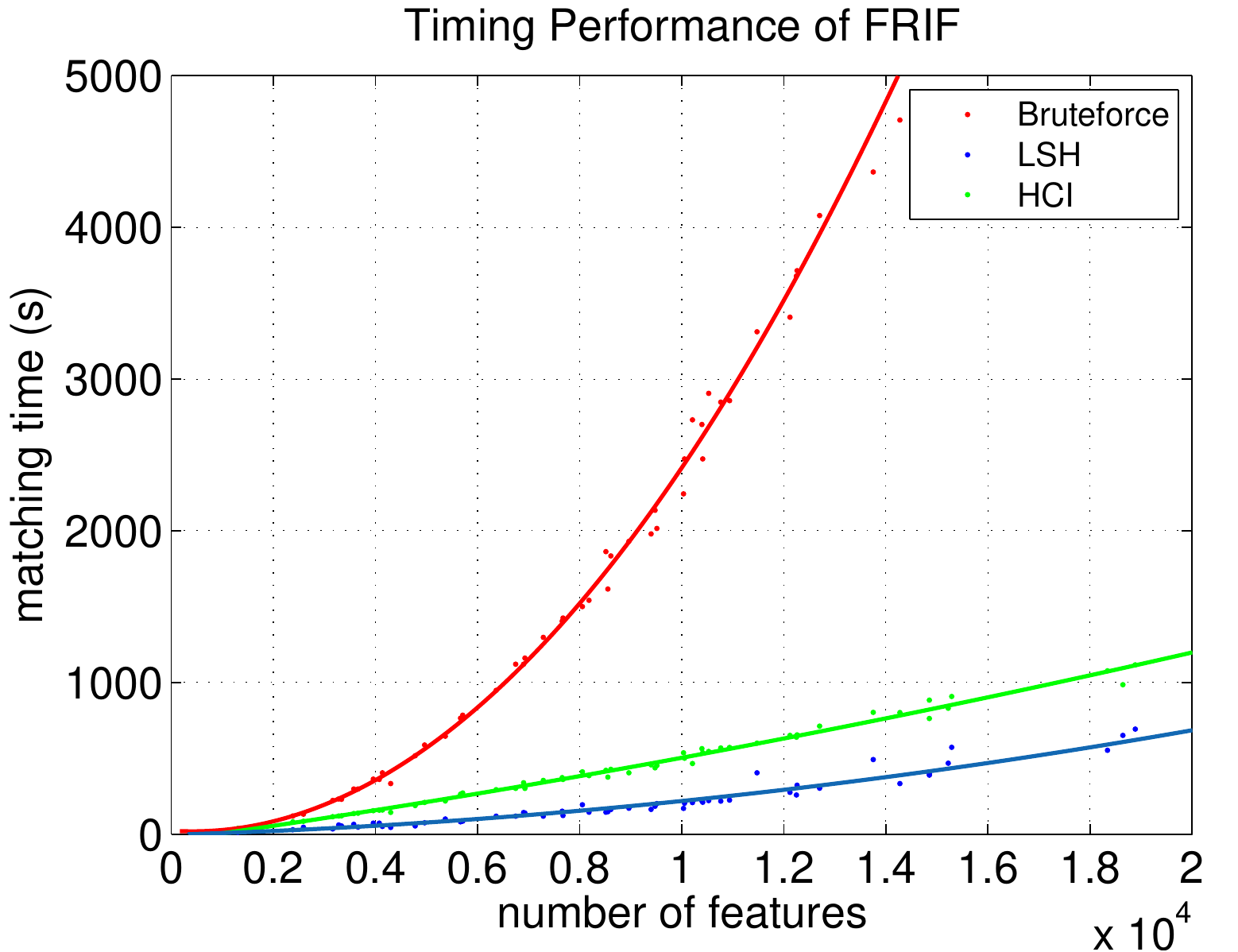}}
	\subfigure[]{
		\includegraphics[width=0.23\textwidth]{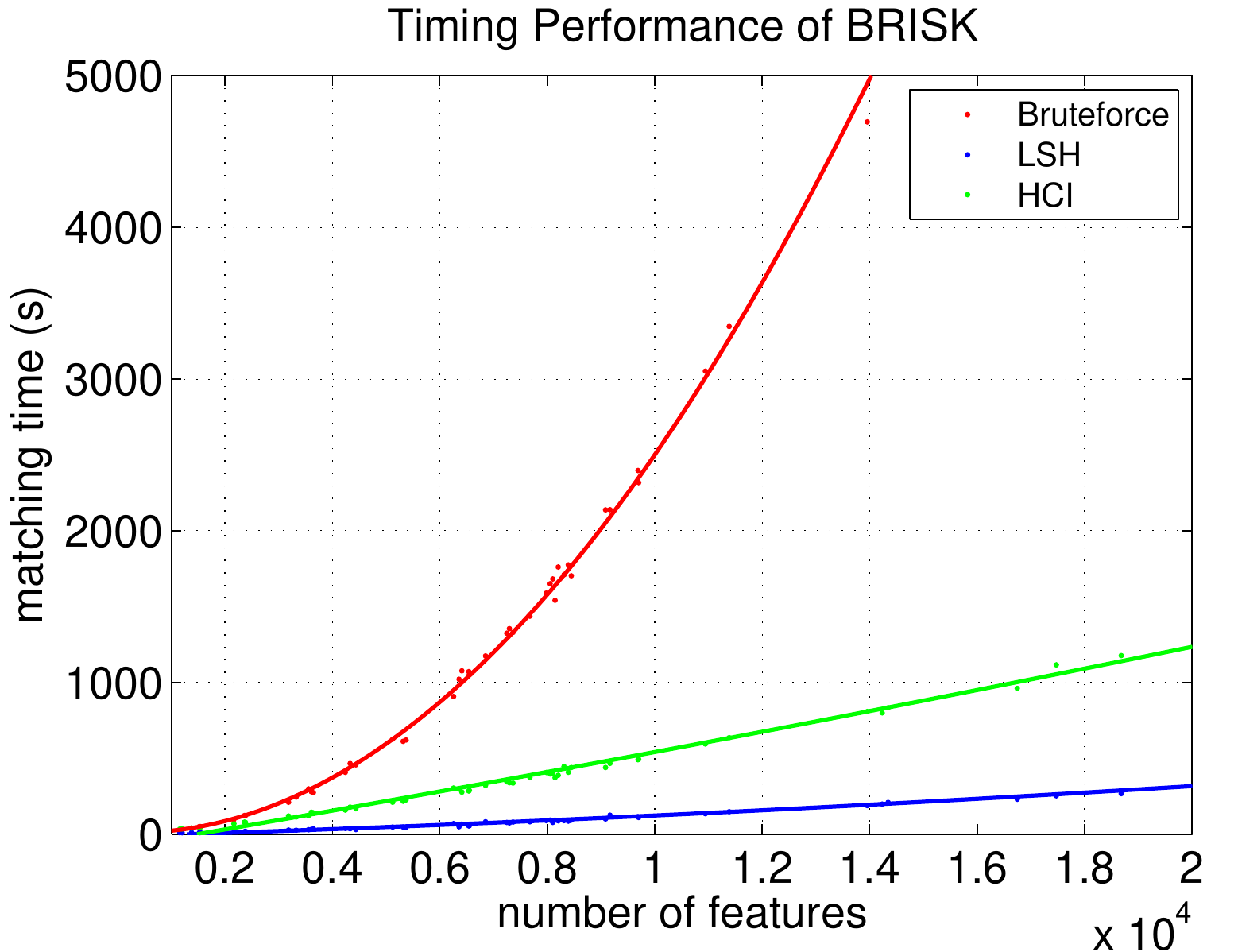}}
	\subfigure[]{
		\includegraphics[width=0.23\textwidth]{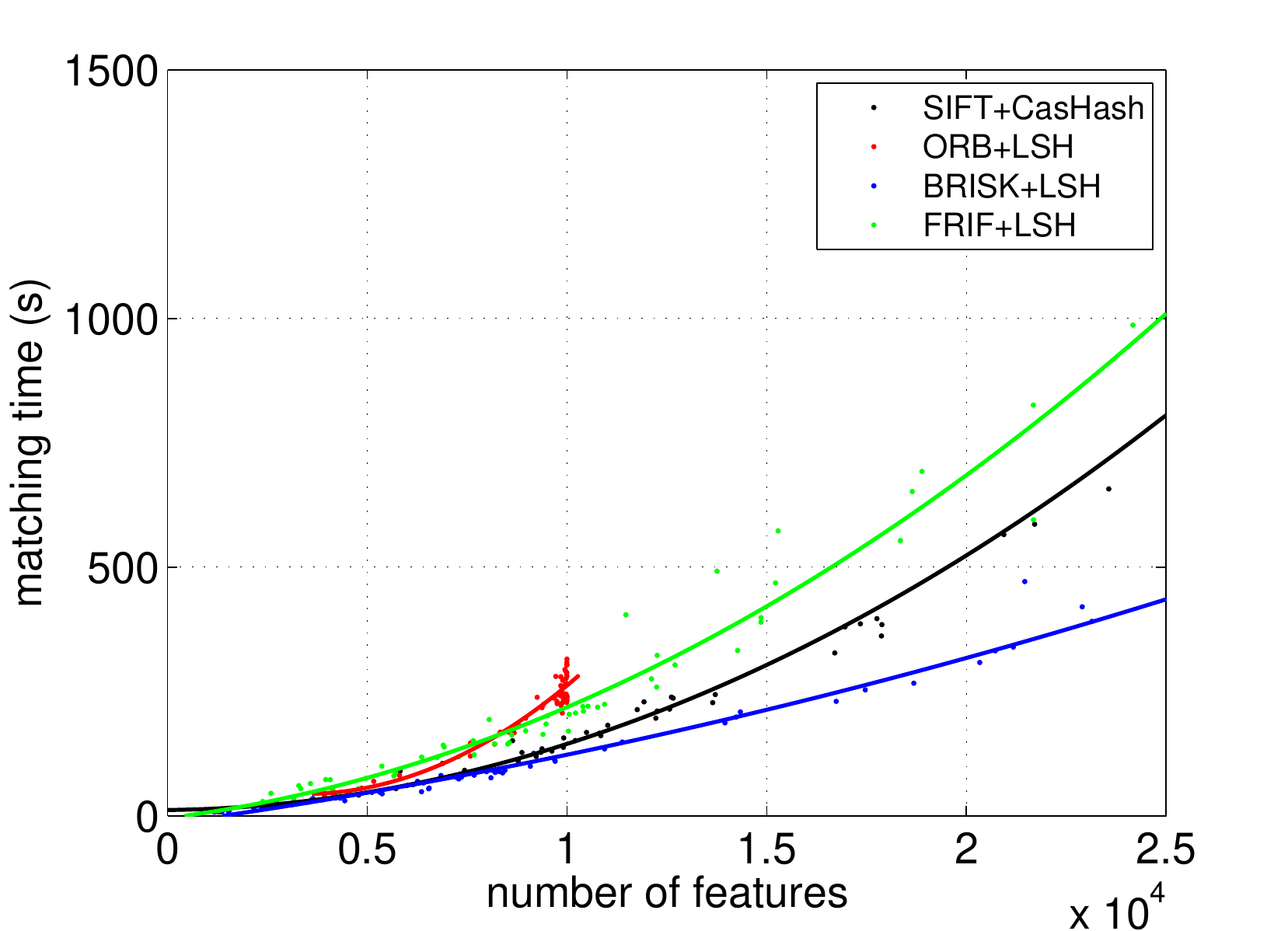}}
	\caption{(a)-(c) are the timing results of different matching methods for ORB, FRIF and BRISK respectively. For a clear comparison among different local features, (d) draws the timing results across different local features. Since LSH is more effective than HCI, binary features matched by LSH are chose to depicted in (d). Note that the recorded times are the total times used for matching all image pairs in a scene, and the number of features is the average number for all images in the scene. \label{fig:timing-result}}
\end{figure}

We first examine the timing performance of different feature matching methods. The feature matching time as a function of the number of features is plotted in Fig.~\ref{fig:timing-result}. Here, the feature matching time is the total time used for matching all image pairs in a scene, and the number of features is averaged over all images in the scene. Therefore, for a given feature matching method~(e.g., ORB+LSH), each scene corresponds to a point in this figure. From Fig.~\ref{fig:timing-result}(a)-(c), it is clear that bruteforce is inefficient even if its basic computation is the most efficient Hamming distance. For a number of $N$ features, the complexity of bruteforce is $O(N^2)$ as it has to linear scan over all features. For ANN methods, since it first quickly selects a small number of candidates and then conducts linear scan among them, its complexity is $O(N^d)$, where $d < 2$ and depends on the used method and binary feature. A good ANN method for a specific binary feature should have $d$ as close to 1 as possible. We can find from the results in Fig.~\ref{fig:timing-result} that LSH is consistently better than HCI for all tested binary features. Meanwhile, LSH is more efficient when applying to BRISK than to FRIF. The reason could be that the binary elements in BRISK are scattered more uniformly in data than those in FRIF. This will lead to a more uniform size distribution of the hash buckets in LSH, which further reduces the query time. Fig.~\ref{fig:timing-result}(d) compares all the evaluated binary features to the baseline SIFT matching. Due to the superior performance of LSH over HCI, only the results with LSH are drawn in Fig.~\ref{fig:timing-result}(d). It is interesting to find that SIFT+CasHash is very competitive in matching time, only inferior to the best BRISK+LSH.

\begin{figure}[t]
	\centering \subfigure[]{
		\includegraphics[width=0.23\textwidth]{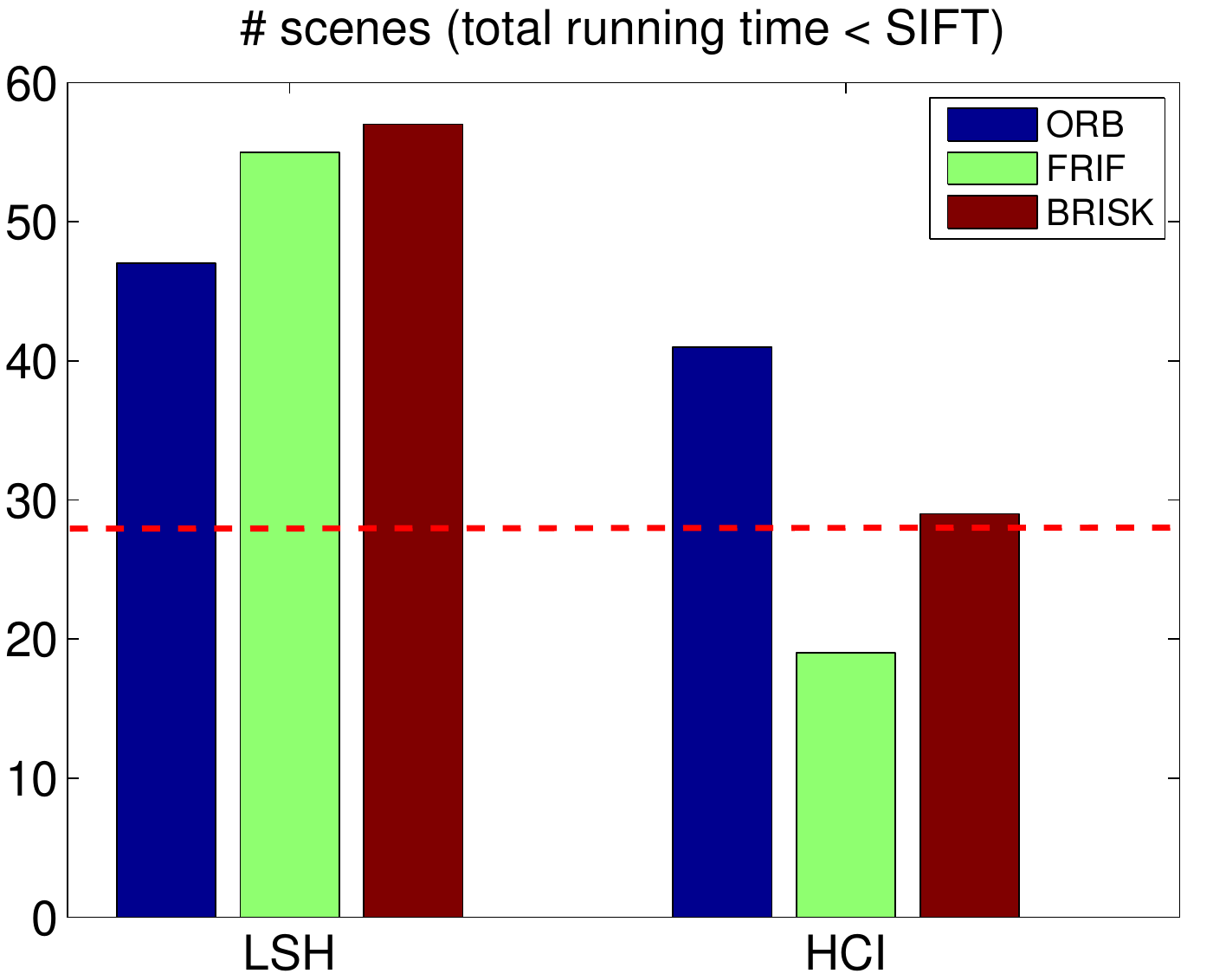}}
	\subfigure[]{
		\includegraphics[width=0.23\textwidth]{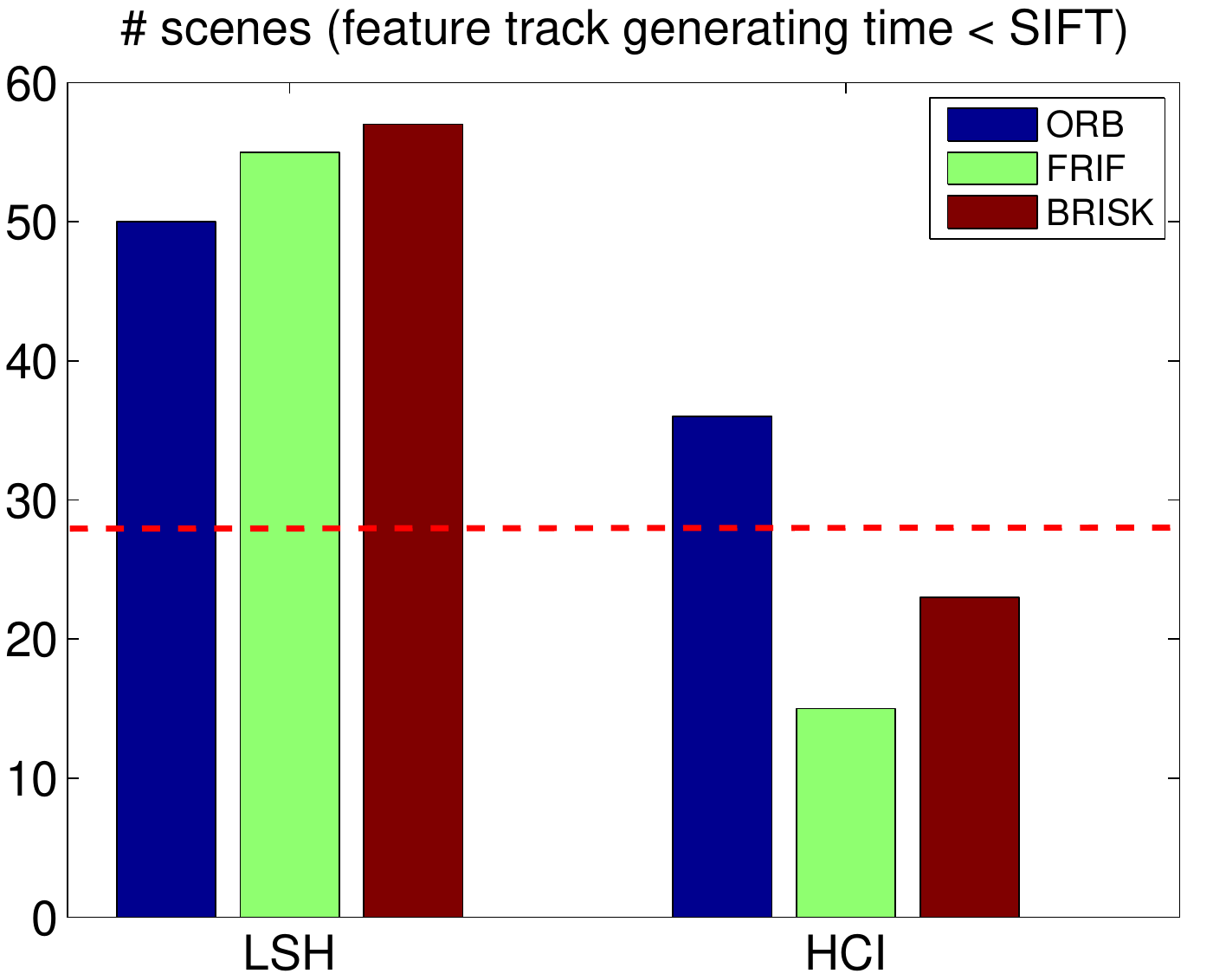}}
	\subfigure[]{
		\includegraphics[width=0.23\textwidth]{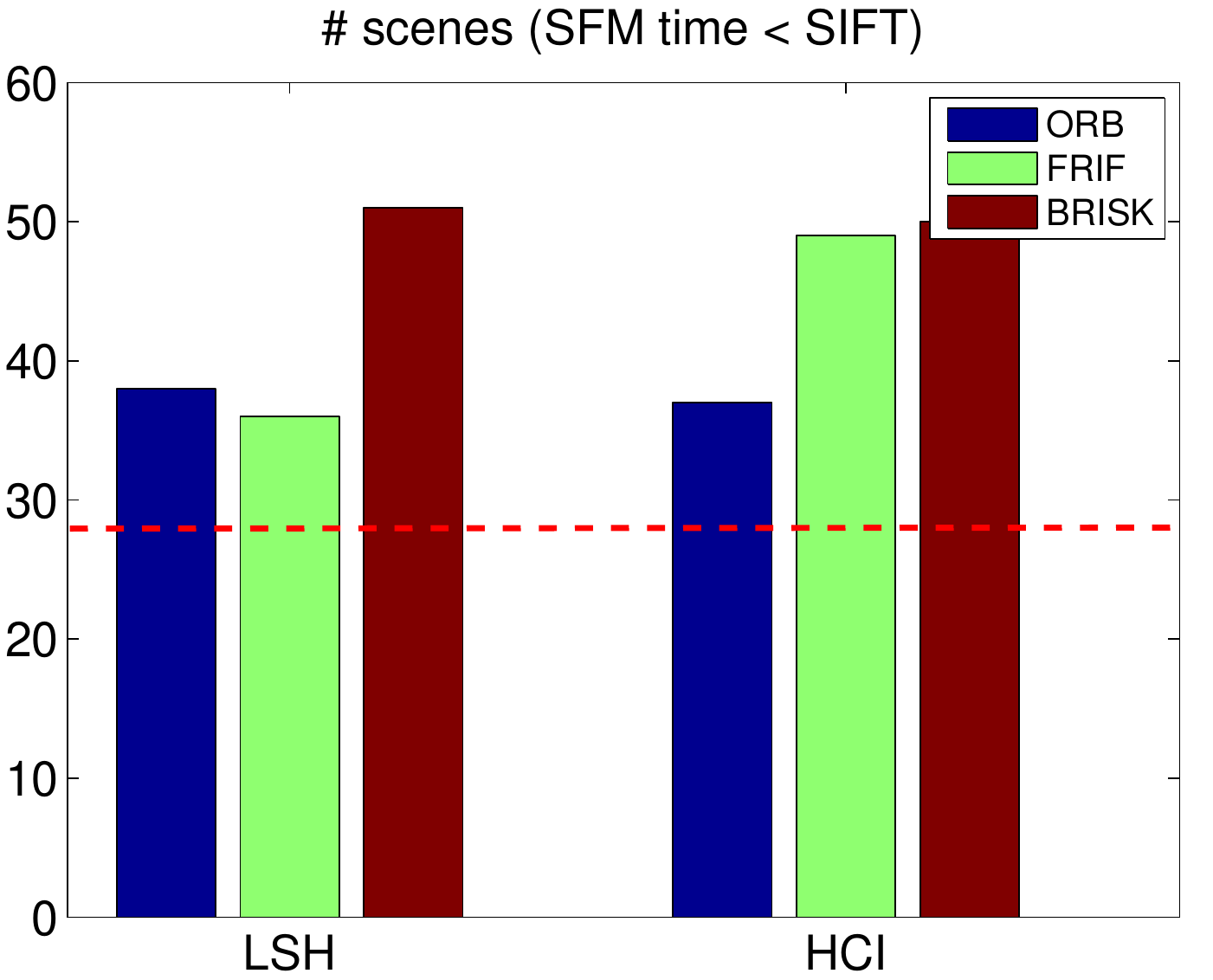}}
	\subfigure[]{
		\includegraphics[width=0.23\textwidth]{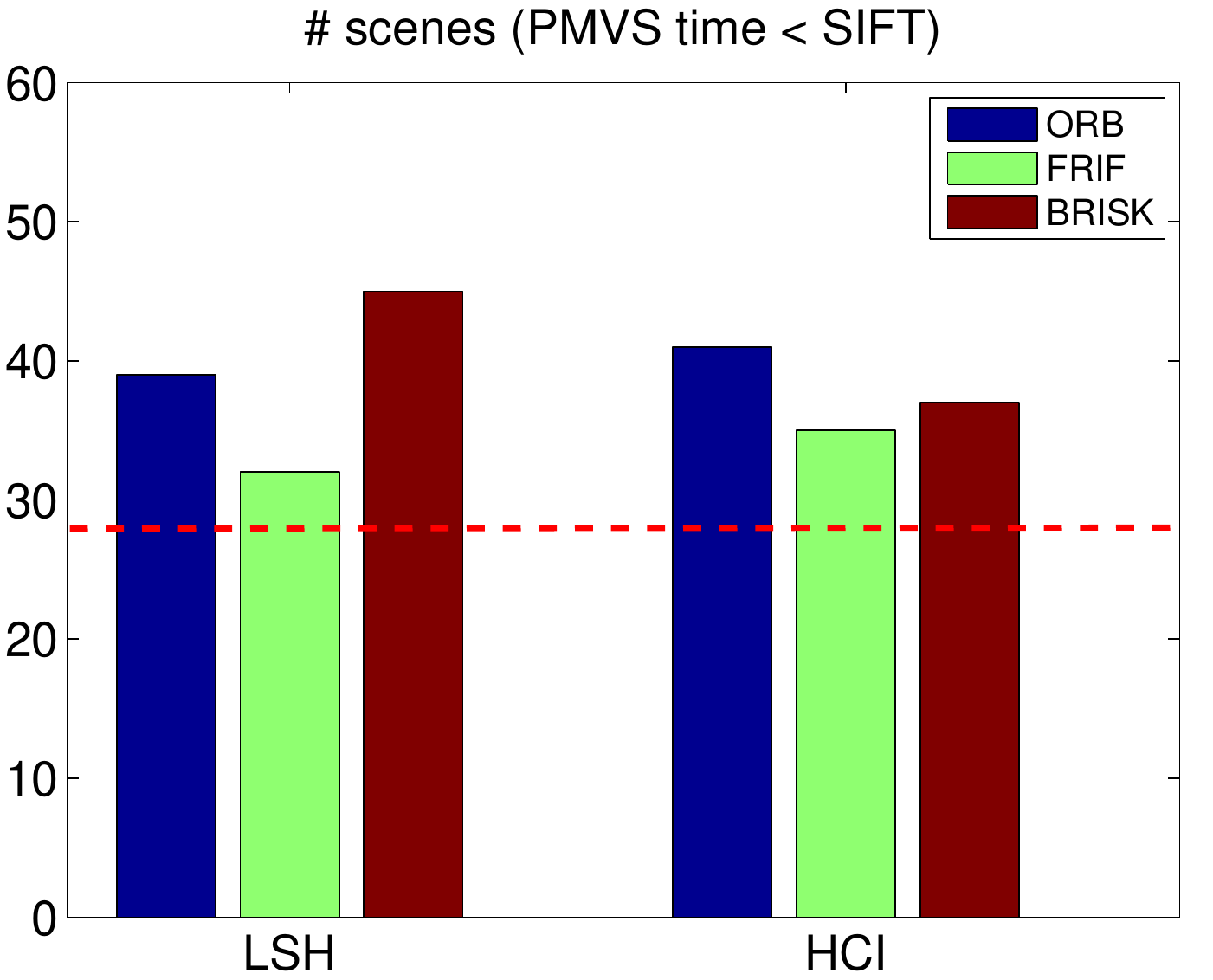}}
	\caption{Running time comparison of different feature matching methods. For a specific feature matching method, e.g., ORB+LSH, the number of scenes for which it takes less time than SIFT+CasHash is given in this figure. (a) shows the total running time of 3D reconstruction, and (b)-(d) show the running times of the three steps of 3D reconstruction. They are, (b) running time of generating feature tracks, (c) running time of SFM, and (d) running time of PMVS. The dash red line indicates the half size of the dataset. A bar higher than its position implies there are more scenes using time less than the baseline method~(SIFT+CasHash). \label{fig:timing-result-2}}
\end{figure}

We then study the total running times for different methods, as well as the timing results in each part of the 3D reconstruction system. Fig.~\ref{fig:timing-result-2}(a) shows the statistic of total running time by using different binary feature matching methods. For a specific method, such as ORB+LSH, its running time is compared to the baseline SIFT+CasHash. We count the number of scenes for which the baseline has more running time and show the results as bars in Fig.~\ref{fig:timing-result-2}. Therefore, a higher bar means there are more scenes that the tested method is more efficient than SIFT+CasHash. As can be seen, the most time efficient method is BRISK+LSH, closely followed by FRIF+LSH. We can also find that when using HCI instead of LSH, both BRISK and FRIF degrade a lot in terms of speed. Although SIFT+CasHash takes less time in matching features than FRIF+LSH and ORB+LSH~(cf. Fig.~\ref{fig:timing-result}(d)), its total running time is higher than theirs because of the fast feature extraction procedures of FRIF and ORB. This is shown in Fig.~\ref{fig:timing-result-2}(b), where the summing time of feature extraction and feature matching of SIFT is higher than those of ORB+LSH and FRIF+LSH. For running time of SFM shown in Fig.~\ref{fig:timing-result-2}(c), all binary methods are faster than SIFT+CasHash. This is due to the fact that the larger number of feature tracks can be generated by SIFT+CasHash. Finally, since there are more recovered cameras obtained by SIFT+CasHash, it is reasonable to spend more time on PMVS than other methods as shown in Fig.~\ref{fig:timing-result-2}(d). Meanwhile, as demonstrated in Fig.~\ref{fig:cam-result}(d), FRIF+LSH performs very close to SIFT+CasHash for SFM, so does it in time usage of PMVS.

%

\section{Conclusion}
In this paper, we conduct a performance evaluation of binary features for 3D reconstruction. We have tested three popular ones~(ORB, FRIF and BRISK) and two related ANN matching methods~(LSH and HCI). Based on the experimental results on a recently proposed 3D reconstruction dataset, we find that FRIF performs the best among these binary features. It turns out that the advantage of using binary features lies in its speed, but the most efficient BRISK achieves this at a large cost of performance degradation. Meanwhile, although with the most efficient Hamming distance, bruteforce matching of binary features is still impractical and requires some kind of ANN methods, for which LSH is consistently better than HCI in our evaluations. Overall, FRIF achieves a satisfactory tradeoff between accuracy and running time, with a slightly worse accuracy and a little faster running time than SIFT.

{\small
\bibliographystyle{ieee}
\bibliography{egbib}
}

\end{document}